\documentclass[a4paper]{cas-sc}
\usepackage[numbers]{natbib}
\usepackage{subcaption}
\usepackage{hyperref}
\usepackage{url}
\usepackage{multirow}
\usepackage{booktabs}
\usepackage{amssymb}
\usepackage{mathrsfs}
\usepackage{amsmath}
\usepackage{bbding}
\usepackage{graphicx}
\PassOptionsToPackage{table}{xcolor}
\usepackage{colortbl}  %彩色表格需要加载的宏包
\usepackage{wrapfig}
\usepackage{subcaption}
\usepackage{floatrow}
\usepackage{pifont}
\usepackage{color}
\usepackage{anyfontsize}
\usepackage[11pt]{moresize}
\usepackage[export]{adjustbox}
\definecolor{defaultcolor}{gray}{0.9}
\usepackage[linewidth=1pt]{mdframed}
\usepackage[normalem]{ulem}
\usepackage{enumitem}

\newcommand{\mat}[1]{\mathbf{#1}}

\definecolor{sh_gray}{rgb}{0.84,0.84,0.84}
\definecolor{sh_gray2}{rgb}{1,0.89,0.75}
\definecolor{color3}{rgb}{0.95,0.95,0.95}
\definecolor{color4}{rgb}{0.94,0.94,1}
\definecolor{color5}{rgb}{1,0.96,0.88}

\hypersetup{   % 设置跳转连接颜色
     colorlinks=true,
     linkcolor=red,
     filecolor=blue,
     citecolor=blue,      
     urlcolor=cyan
     }
\def\tsc#1{\csdef{#1}{\textsc{\lowercase{#1}}\xspace}}
\tsc{WGM}
\tsc{QE}
\begin{document}
\let\WriteBookmarks\relax
\def\floatpagepagefraction{1}
\def\textpagefraction{.001}
\let\printorcid\relax % 可去掉页面下方的ORCID(s)

% \shorttitle{<short title of the paper for running head>} 
\shorttitle{}    

% \shortauthors{<short author list for running head>}
\shortauthors{Ming Xiao et al.}

%论文标题
\title[mode = title]{Defense against Unauthorized Distillation in Image Restoration via Feature Space Perturbation}  

%作者信息
\author[1]{Han Hu}
\author[2]{Zhuoran Zheng}
\author[1]{Chen Lyu}

\cormark[1]

\address[1]{Shandong Normal University, Jinan, China} %声明第一单位
\address[2]{Sun Yat-sen University, Guangzhou, China} %声明第二单位

\cortext[3]{Corresponding author}  %声明通讯作者

\address[]{huhan199908@163.com, zhengzr@njust.edu.cn,  lvchen@sdnu.edu.cn}

\begin{abstract}
Knowledge distillation (KD) attacks pose a significant threat to deep model intellectual property by enabling adversaries to train student networks using a teacher model’s outputs. While recent defenses in image classification have successfully disrupted KD by perturbing output probabilities, extending these methods to image restoration is difficult. Unlike classification, restoration is a generative task with continuous, high-dimensional outputs that depend on spatial coherence and fine details. Minor perturbations are often insufficient, as students can still learn the underlying mapping. 
% Indeed, existing undistillation methods designed for classification  which typically inject noise into output logits  largely fail in restoration tasks because they do not interfere with the rich intermediate feature representations that students can exploit.
%
To address this, we propose Adaptive Singular Value Perturbation (ASVP), a runtime defense tailored for image restoration models. ASVP operates on internal feature maps of the teacher using singular value decomposition (SVD). It amplifies the top-k singular values to inject structured, high-frequency perturbations, disrupting the alignment needed for distillation. This hinders student learning while preserving the teacher's output quality.
We evaluate ASVP across five image restoration tasks: super-resolution, low-light enhancement, underwater enhancement, dehazing, and deraining. Experiments show ASVP reduces student PSNR by up to 4 dB and SSIM by 60–75\%, with negligible impact on the teacher’s performance. Compared to prior methods, ASVP offers a stronger and more consistent defense.Our approach provides a practical solution to protect open-source restoration models from unauthorized knowledge distillation.
\end{abstract}


% \begin{highlights}
% \item highlight-1
% \item highlight-2
% \item highlight-3
% \end{highlights}

% Keywords
\begin{keywords}
Knowledge distillation \sep
Image Restoration Security \sep
Singular Value Perturbation \sep 
Undistillationn Defense \sep 

\end{keywords}

\maketitle
\section{Introduction}
 Image restoration is a fundamental task in computer vision that aims to reconstruct high quality images from degraded inputs. In recent years, deep learning based methods have demonstrated remarkable success across a wide range of restoration tasks such as medical imaging, remote sensing, and cultural heritage preservation owing to their powerful feature representation capabilities. To enable real time deployment on resource constrained devices, knowledge distillation (KD) \cite{hinton2015distilling_1} has been widely adopted to compress large, high performance teacher models into lightweight student networks, significantly reducing computational overhead while preserving reconstruction quality. Previous studies have successfully applied KD to various image restoration domains, including super resolution\cite{li2023diffusion2},\cite{hui2018fast3},\cite{gao2018image4}, denoising\cite{zhuo2019ridnet5},\cite{li2022multiple6},\cite{lin2022lightweight7}, deraining\cite{cui2022semi8},\cite{luo2023local9},\cite{li2025sequence10}, and medical image reconstruction\cite{murugesan2020kd11}, by using the teacher's outputs or intermediate features as ``soft labels`` to guide student training. However, this widespread adoption of KD also introduces a serious security concern: adversaries may exploit the distillation process to steal a model’s knowledge (e.g., through model extraction attacks) \cite{lopes2017data12},\cite{zhang2021data13},\cite{chawla2021data14}, posing a direct threat to the intellectual property (IP) of the original model developers\cite{gou2021knowledge15},\cite{liu2020residual16},\cite{quan2022self17}.

\begin{figure}[!tb]
\centering

\setlength{\tabcolsep}{1pt}
\begin{adjustbox}{width=0.8\textwidth}
\renewcommand{\arraystretch}{1.2}
\begin{tabular}{>{\centering\arraybackslash}m{1.5cm} 
                >{\centering\arraybackslash}m{2.5cm} 
                >{\centering\arraybackslash}m{2.5cm} 
                >{\centering\arraybackslash}m{2.5cm} 
                >{\centering\arraybackslash}m{2.5cm}}
                
 & normal & $\beta$: low & $\beta$: high  & ours \\

\multirow{2}{*}{$T-\beta$} 
 & \includegraphics[width=2.5cm]{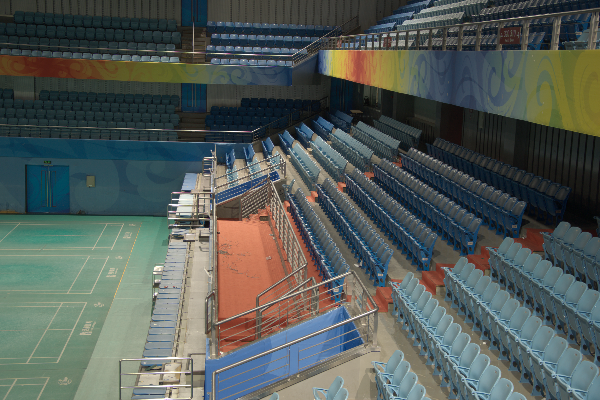}
& \includegraphics[width=2.5cm]{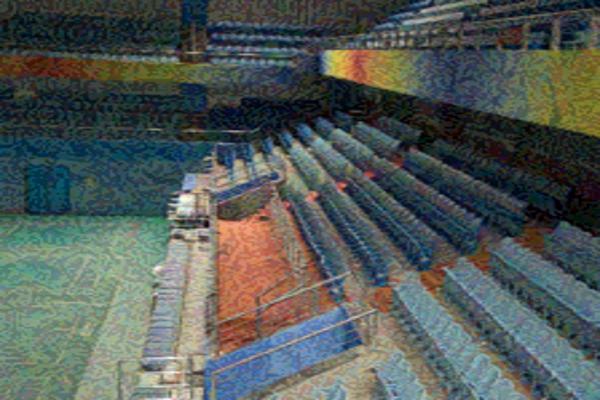}
& \includegraphics[width=2.5cm]{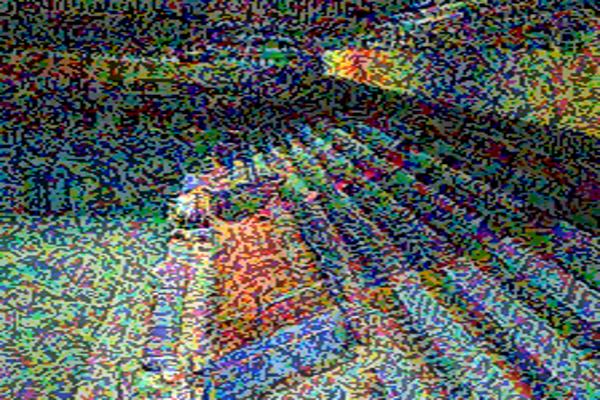}
& \includegraphics[width=2.5cm]{introduction/gt.png} \\

% & \rowcolor{blue!10}
SSIM & \textcolor{red}{0.644} & 0.617 & 0.327 & 0.644 \\

\multirow{2}{*}{$S-KD$} 
& \includegraphics[width=2.5cm]{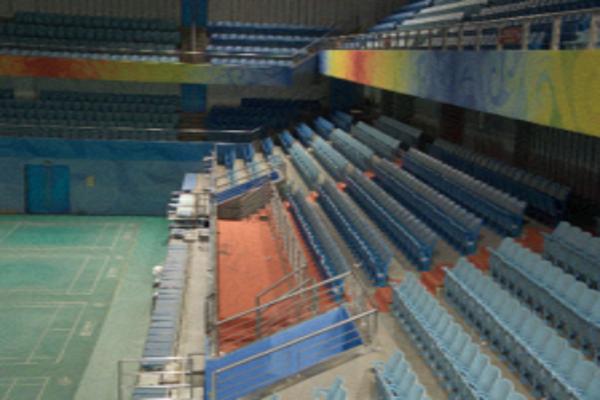}
& \includegraphics[width=2.5cm]{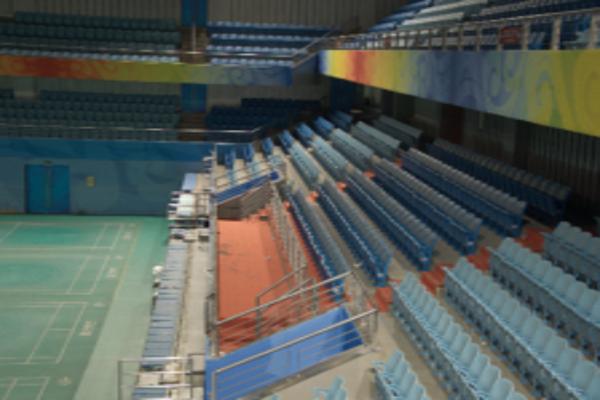}
& \includegraphics[width=2.5cm]{introduction/kd.jpg}
& \includegraphics[width=2.5cm]{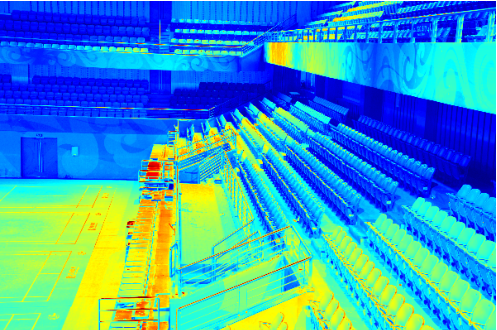} \\

% & \rowcolor{blue!10}
SSIM & \textcolor{red}{0.676} & 0.668 & 0.619 & 0.197 \\

\end{tabular}
\end{adjustbox}
\caption{Visual comparison under different $\beta$ values and corresponding SSIM scores for $T-\beta$ (teacher model) and $S-KD$ (distilled student model) on a low light enhancement task. This experiment explores the transfer of adversarial perturbation strategies from classification to restoration. As $\beta$ increases, the teacher's output degrades significantly due to amplified noise, whereas the student maintains relatively stable performance. This highlights the challenge of applying classification based defenses to generative tasks like image enhancement.}
\label{fig:introduction}
\end{figure}

In the classification domain, several defensive strategies have been proposed to counter unauthorized distillation attacks such as the ``stingy teacher``\cite{ma2021undistillable19} and ``nasty teacher``\cite{ma2022stingy20} approaches. These typically work by perturbing the teacher’s outputs (e.g., through noise injection or output sparsification) to mislead the student, while preserving the teacher’s classification accuracy. For example, the teacher model can be trained to output misleading probability distributions that confuse the student without affecting its own predictions. However, such output level defenses are largely ineffective when directly applied to image restoration. Unlike classification, which deals with discrete probability vectors, image restoration generates continuous, high resolution images.
 Even if the outputs are slightly perturbed, student models can still extract useful information.

To address this issue, we firstly explored a variety of feature space interventions to protect restoration models. Specifically, we experimented with injecting Gaussian noise into intermediate features, randomly dropping feature channels, and applying adversarial perturbations. Our findings revealed a critical trade off: low intensity perturbations are too weak to stop knowledge transfer, while strong perturbations can suppress student performance but severely harm the teacher’s restoration ability. Excessive feature distortion disrupts the global structure and fine details of the teacher’s outputs, leading to significant performance drops. These observations highlight the limitations of naïve feature space defenses: gentle perturbations are ineffective, and aggressive ones compromise what we seek to protect.

To address the above challenge, we propose a runtime feature-space defense module, termed Adaptive Singular Value Perturbation (ASVP). Without requiring retraining or architectural modifications, ASVP effectively protects the teacher model during inference. The core idea is to perform Singular Value Decomposition (SVD) on each intermediate feature map of the teacher model and amplify the top-$k$ singular values, thereby introducing structured high-frequency perturbations. These perturbations disrupt the student model’s learning process by forcing it to follow unstable signals, resulting in inaccurate or noisy representations.
As shown in Fig.~\ref{fig:introduction}, adversarial sample injection methods can negatively affect the performance of the teacher model, especially under high-intensity injections where the degradation becomes more pronounced. Meanwhile, the student model distilled by the attacker can still achieve performance close to that of an unprotected teacher. In contrast, with the proposed ASVP defense, the teacher model maintains its original performance while effectively preventing the attacker from acquiring its knowledge, thus achieving stronger security guarantees. 

Our key contributions are summarized as follows:

\begin{itemize}
    \item[$\bullet$]\textbf{New defense paradigm.} We propose a dynamic feature space framework that injects targeted spectral perturbations by amplifying dominant singular values in real time, effectively preventing student models from aligning with the teacher’s internal features.
    \item[$\bullet$]\textbf{Plug and play.} ASVP is lightweight and easy to integrate into existing architectures. It requires no additional training or modifications, making it deployable even on already trained models while effectively encrypting the feature space.
    \item[$\bullet$]\textbf{Extensive empirical validation.} We evaluate our approach across five representative image restoration tasks. Results show that ASVP significantly reduces student performance (e.g., up to 4 dB drop in PSNR and 60–75$\%$ drop in SSIM) with minimal impact on the teacher’s output quality. To the best of our knowledge, this is the first defense that achieves such effectiveness in restoration tasks without sacrificing fidelity.
\end{itemize}

\section{Related work}
\subsection{Knowledge Distillation for Image Restoration}
KD has been widely adopted in image restoration as an effective technique for compressing large teacher models into lightweight student networks, enabling real time deployment without significant loss in restoration quality~\cite{murugesan2020kd-MRI111}. Since its introduction into the restoration domain, KD has evolved across several key strategies.

Early works mainly focused on output level distillation, where the student minimizes the mean squared error (MSE) between its output and that of the teacher. For example, Gao et al.~\cite{gao2018image4} trained student networks using teacher generated super resolution outputs as soft labels. In image denoising, Li et al.~\cite{li2023diffusion2} proposed residual based distillation, where students mimic the noise maps predicted by the teacher, thus improving convergence and denoising quality.

Building on these foundations, researchers began exploring feature level distillation. Meng et al.~\cite{meng2021gradient2-3} introduced intermediate feature alignment in real time SR, significantly improving detail preservation. Zhou et al.~\cite{zhou2025dynamic2-4} proposed a fidelity preserving framework combining output and multi scale feature supervision. Attention based distillation methods were also developed. Wang et al.~\cite{wang2022panini2-5} used teacher attention maps to guide students toward salient image regions, improving performance in deblurring tasks.

More recent advances have focused on unified and adaptive distillation frameworks. Online joint training frameworks~\cite{lan2022online2-7},\cite{habib2023knowledge_B4},\cite{xu2022unsupervised_B5} integrated data driven and model driven priors for end-to-end restoration, particularly in dehazing. In ~\cite{yu2024towards2-8}, Yu et al. introduced progressive distillation with multi stage compression to balance efficiency and quality. Zhang and Yan~\cite{zhang2025soft2-9} designed a soft distillation scheme leveraging multi dimensional cross network attention and contrastive learning to enhance feature correlation transfer while maintaining strong restoration capability~\cite{zhu2018image-image2222}.

However, as KD continues to permeate restoration pipelines, concerns have emerged regarding unauthorized knowledge extraction and the potential compromise of proprietary or sensitive model IP.

\subsection{Undistillation Defense for IP Protection}

Unlike prior SVD-based distillation methods that employ singular value decomposition to compress or align features for better knowledge transfer, our approach repurposes SVD as a defensive mechanism. Specifically, ASVP amplifies the top-k singular values of teacher features at runtime to inject structured perturbations that disrupt student alignment while preserving the teacher’s fidelity. This fundamentally differs from existing SVD-KD works that aim to facilitate knowledge transfer, whereas our goal is to frustrate unauthorized distillation in image restoration.
To combat knowledge theft via distillation, recent works have begun to explore defenses tailored to image restoration~\cite{alkhulaifi2021knowledge_B2,csimcsek2024screen_B6}. Inspired by adversarial teacher strategies from classification tasks~\cite{surianarayanan2023survey_B1}, these methods seek to degrade student performance without affecting the teacher. However, restoration differs fundamentally from classification—it requires fine spatial coherence and pixel level fidelity. Consequently, simple output perturbations often fail, and stronger defenses risk compromising visual quality.

General IP protection techniques such as digital watermarking~\cite{regazzoni2021protecting_B12,fan2019rethinking_B11} and model passports~\cite{fan2019digital_B10,yao2019latent_B13} can detect theft but cannot prevent functionality replication. Backdoor based methods~\cite{gou2021knowledge15,perry2000digital_B9,picard2004towards_B7} poison the model via triggers but reduce utility and lack generalization. More advanced solutions like phase hiding~\cite{deng2023pirnet2-10} and adversarial encryption~\cite{liu2024recoverable2-11} aim to preserve data privacy rather than model ownership. These methods often assume insecure input or transmission, but do not address student imitation of \emph{internal feature representations}, a key vulnerability in distillation attacks~\cite{herrigel2000optical_B8},\cite{hu2022teacher_B3},\cite{li2021anti_B16},\cite{wang2020backdoor_B14}. Recently, emerging research in the field of generative models has explored the embedding of invisible or copyright-protective information within neural representations. For example, methods such as StegaNeRF~\cite{li2023steganerf} and InstantSPLAMP~\cite{li2025instantsplamp} hide information in neural radiance fields and Gaussian splatting frameworks, aiming to preserve ownership through invisible steganographic cues. Similarly, ConcealGS~\cite{yang2025concealgs} and Hide-in-Motion~\cite{liu2025hide} propose mechanisms for embedding invisible watermarks into 3D and 4D Gaussian assets, respectively. While these approaches are designed for copyright tracking and media protection, they share conceptual similarities with our work in terms of operating directly within intermediate feature spaces. However, unlike steganographic works that preserve model output and embed ownership metadata, our method actively degrades student model alignment by injecting perturbations during feature distillation, focusing on defense against unauthorized imitation rather than passive traceability. This contrast highlights the novel application of structured perturbation as an active IP protection strategy in vision restoration pipelines. In addition, works like Generator versus Segmentor~\cite{zhang2021generator} explore pseudo-healthy image synthesis by aligning generative and segmentation objectives, offering novel insights into task-specific representation manipulation. Although their goal is to reconstruct clean anatomical structures rather than defend feature access, such methods share a high-level intent of controlling internal feature behavior, further motivating the need for robust defenses like ASVP in restoration pipelines.

In summary, current defenses either sacrifice output quality or leave intermediate features unprotected. To address this, we propose a new direction: injecting high frequency, targeted perturbations into the feature space at runtime via Adaptive Singular Value Perturbation. By modifying internal structure rather than output, this strategy prevents student alignment while preserving teacher fidelity, offering a practical and scalable solution for protecting open source restoration models against unauthorized KD.

\begin{figure*}[!htp]
    \centering
    \includegraphics[width=1\linewidth]{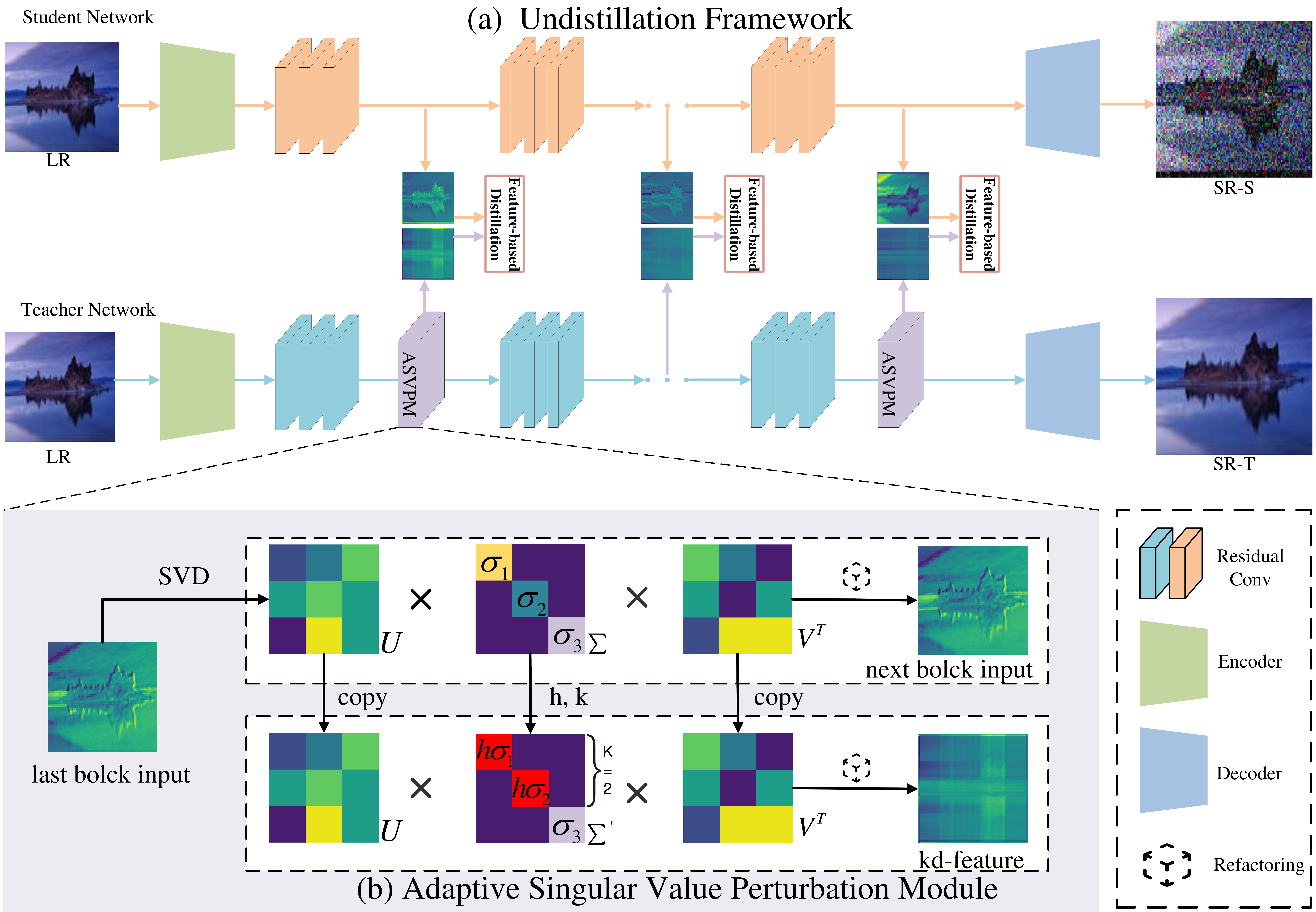}
    \caption{Overview of the proposed framework. (a) Standard distillation where the student mimics intermediate features from the teacher. (b) ASVP perturbs features by amplifying the top-$k$ singular values with factor $h$ via SVD to block distillation.}

    \label{fig:framework}
\end{figure*}

\section{Method}
This section details the principle and implementation of the proposed ASVP) module. We begin by outlining the overall undistillation framework for image restoration, then describe the design and operation of the ASVP module. Finally, we present a theoretical analysis of the method, including its impact on feature information transfer, computational complexity, and the dual path design that preserves the teacher’s performance.
\subsection{Framework}
To protect an image restoration model against knowledge distillation attacks, we introduce a lightweight runtime defense that operates directly on the teacher’s intermediate feature maps. As illustrated in Fig.~\ref{fig:framework}, the framework comprises two main components: (a) a standard knowledge distillation pipeline in which ASVP modules are inserted into the teacher model between residual blocks, and (b) the internal mechanism of the ASVP modules itself, which perturbs feature representations in a controlled manner to produce “encrypted” features for the student.

In our distillation setting (Fig.~\ref{fig:framework}(a)), the teacher and student networks share a typical encoder residual decoder architecture and process the same input (e.g., a low resolution image in a super resolution task). During distillation training, the student attempts to align its intermediate features with those of the teacher. However, the teacher’s features are intercepted and transformed by ASVP modules before the student can access them. Each ASVP module operates on the fly during inference, sitting between two residual blocks of the teacher. The teacher’s final output image (denoted SR-T for super resolution teacher output) remains clean and high quality, since the perturbations are not propagated through the teacher’s own forward path. In contrast, the student, which learns from the perturbed intermediate features, produces degraded results (SR-S for student output) due to its inability to mimic the unstable teacher features. This defense requires no retraining of the teacher model and adds only minimal computational overhead, making it easily applicable to standard restoration architectures without sacrificing the teacher’s performance.

\subsection{Adaptive Singular Value Perturbation Module}
The ASVP module is the core component of our defense framework, designed to safeguard the teacher’s intermediate representations. Its primary goal is to prevent the student model from directly accessing the teacher’s original feature maps, thereby reducing the risk of knowledge leakage. Fig.~\ref{fig:framework}(b) provides an overview of the module’s operation. At a high level, the ASVP module performs a spectral perturbation of the feature map using singular value decomposition and selective amplification.

\subsubsection{Singular Value Decomposition}Specifically, the module takes the output feature map from the previous network block and performs a Singular Value Decomposition (SVD) on it. This yields:

\begin{equation}
\mat{X}=\mat{U}\mat{\Sigma^{\prime}} \mat{V}^\top,
\end{equation}

% \noindent where $\mat{U}$ and $\mat{V}$ are orthogonal matrices, and $\mat{\Sigma}$ is a diagonal matrix containing singular values $\sigma_1 \ge \sigma_2 \ge \cdots \ge \sigma_r$, which represent the directional importance of the feature content.

\noindent  where $\mat{U}$ and $\mat{V}$ are orthogonal matrices, and $\mat{\Sigma}$ is a diagonal matrix whose diagonal elements, known as singular values($\sigma_1, \sigma_2, \cdots,\sigma_r$), are arranged in non-increasing order $(\sigma_1 > \sigma_2 > \cdots > \sigma_r)$, These values quantify the relative significance of different directions in the feature space.

\subsubsection{Singular Value Perturbation\textbf{(SVP)}}The top-$k$ singular values are scaled by an amplification factor $h$, controlled by hyperparameters $k$ and $h$. The modified singular value matrix $\mat{\Sigma}^{\prime}$ is constructed as:

\begin{equation}
\mat{\Sigma^{\prime}}=\begin{cases}h\cdot\sigma_j&j\leq k\\
\sigma_j&j>k\end{cases}
\end{equation}

\noindent $\sigma_1, \dots, \sigma_k$ are each multiplied by $h$, while the remaining singular values are left unchanged. By amplifying these top-$k$ singular values, the module injects a structured perturbation concentrating on the most informative directions of the feature. This selective amplification dramatically alters the feature content along those principal components, creating an unstable or noisy version of the feature map that is difficult for the student to learn. Crucially, this perturbation is structured (non-random) and focuses on high variance feature components, as opposed to adding arbitrary noise this ensures that the distortion effectively targets the “teachability” of the features without simply destroying them indiscriminately.

%This selective amplification perturbs the most informative components of the feature map, injecting structured %disruption that significantly impairs the student’s ability to extract meaningful knowledge.

\subsubsection{Feature Reconstruction and Integration}Using the amplified singular values $\mat{\Sigma}^{\prime}$, the encrypted feature map is reconstructed:

\begin{equation}
    \mat{X}^{\prime}=\mat{U}\mat{\Sigma}^{\prime} \mat{V}^\top
\end{equation}

This encrypted feature map $\mat{X}^{\prime}$ is then provided to the student. Meanwhile, the original (unperturbed) feature $\mat{X}$ is passed unchanged to the next block in the teacher network. This ensures that the teacher continues normal inference without degradation, while the student learns from perturbed signals. The ASVP module thus operates as a plug and play unit that encrypts only the distillation path, preserving performance on the teacher side.

\subsubsection{Novelty \& Design Rationale}
Knowledge distillation attacks succeed in image restoration because students can align with the teacher’s internal signals even when output noise is added. To counter this, we defend inside the feature space at runtime. Our module, ASVP, is a lightweight component inserted between residual blocks. At each forward pass it decouples the representation into two paths: a clean path that continues the teacher’s normal inference, and a protected path exposed only to the student. On the protected path, ASVP performs a spectral transformation of the intermediate feature map—amplifying its top-k singular values—to inject structured, high-frequency perturbations that deliberately make feature alignment unstable for the student while keeping the teacher’s output unaffected. The design is model-agnostic, requires no retraining or architectural changes, and is meant to be dropped into existing restoration pipelines with modest overhead.We emphasize that the dual-path design of ASVP ensures strict isolation between the teacher's clean inference path and the perturbed features exposed to the student. The spectral perturbation is applied to a copied feature map, and no backward or forward flow exists from the perturbed branch into the teacher's main computation graph. This prevents any interference with residual connections or batch normalization statistics, preserving the integrity of the teacher’s inference.
\\ To make our contributions explicit, we then highlight three core innovations of ASVP:
\begin{enumerate}[label=\Roman*] 
\item Plug-and-play runtime defense with a dual-path design.
Deployed at inference between residual blocks; the student receives only the perturbed features while the teacher proceeds on a clean path, preserving teacher fidelity without any retraining. 
\item Structured spectral perturbation via top-k amplification.
Rather than random noise, ASVP amplifies the top-k singular values to inject targeted, principal-direction perturbations that specifically undermine student alignment yet are neutralized along the teacher’s clean path.
\item Consistent cross-task effectiveness with practical overhead.
Across diverse restoration tasks (super-resolution, low-light, underwater enhancement, dehazing, deraining), ASVP keeps the teacher’s quality intact while reliably degrading the student, and the added computation is confined to inference and remains modest for typical feature sizes.
\end{enumerate}

\begin{comment}
\begin{algorithm}[!htp]
\caption{Overall process of ASVP}
\label{alg:SVMPM}

\textbf{Input:}{$\mat{X}$ is last residual block output}\\
\textbf{Output:}{$\mat{X}$ is next residual block input, 
                 $\mat{X}^{\prime}$ is KD-feature}

\textbf{Start:}
\begin{enumerate}

    \item \( \mat{U} \mat{\Sigma} \mat{V}^\top \xleftarrow{SVD} \mat{X} \)
    \vspace{4pt}
  
    % \item \( h \sigma_1  \dots, h \sigma_k\dots, h \sigma_r \xleftarrow{\textbf{h,k}}  \sigma_1 \dots, \sigma_k\dots,\sigma_r \\
    %             \Sigma' \xleftarrow{\textbf{h,k}} \Sigma \)
    
    \item \(  \mat{X}^\prime \xleftarrow[]{SVP}\mat{X}: \mat{\Sigma}^\prime   \xleftarrow{h,k} \mat{\Sigma}   \)
    \vspace{4pt}

    \item \( \mat{X}^{\prime} \xleftarrow{FRI} \mat{U} \mat{\Sigma}^{\prime} \mat{V}^{\top} \)
    \vspace{4pt}
    \item \( \mat{X} \xleftarrow{{FRI}} \mat{U} \mat{\Sigma} \mat{V}^{\top} \)
    
\end{enumerate}

\textbf{end}
\end{algorithm}
\end{comment}

\subsection{Analytical Study of the Algorithm}
For deeper insight into the defensive efficacy of ASVP, we analyze its theoretical properties. In particular, we examine: the extent of feature information distortion introduced, the computational overhead of the method, the effect of the dual path strategy on network stability.

\subsubsection{Feature Perturbation Magnitude}By analyzing the Frobenius norm of the difference between pre  and post perturbation feature maps, the perturbation energy is given by:

\begin{equation}
\|\mat{X_i}^{\prime}-\mat{X_i}\|_F^2=\sum_{j=1}^k(h-1)^2\sigma_j^2,
\end{equation}

This shows that the perturbation strength increases quadratically with $h$ and linearly with the sum of squared top-$k$ singular values. The injected perturbation targets the most representative directions, maximizing the difficulty for the student to learn. Accordingly, the injected energy can be expressed as:

\begin{equation}
\Delta_E=(h - 1)^2\cdot\sum_{j=1}^k\sigma_j^2.
\end{equation}

This result quantitatively illustrates the extent to which the ASVP module perturbs the structural integrity of principal features. In essence, ASVP introduces structured noise along the directions of the most significant singular vectors, thereby severely impeding the student model's ability to capture and learn these high value representations.

\subsubsection{Complexity Analysis}Let the feature map dimensions be $m = H \cdot W$ and $n = C$. The computational complexity of performing singular value decomposition (SVD) on an $m \times n$ matrix is:

\begin{equation}
\mathcal{O}(B\cdot\min(mn^2,m^2n)),
\end{equation}

In practice, intermediate feature maps in image restoration networks typically have moderate dimensions (e.g., $64 \times 64 \times 64$ yielding $m, n \approx 10^4$), and the SVD is executed only during inference. As such, the computational overhead introduced by ASVP remains within acceptable limits for real world deployment.

In terms of memory consumption, ASVP requires only temporary storage of the SVD outputs: the left singular vectors $\mat{U}$, the right singular vectors $\mat{V}^\top$, and the modified diagonal matrix $\mat{\Sigma}^\prime$. The original feature map $\mat{X}$ is immediately forwarded to the teacher model’s subsequent layers, avoiding duplication. In summary, ASVP introduces a negligible computational and memory burden per forward pass, which is a worthwhile trade off for the enhanced model protection it provides.

\subsubsection{Dual Path Robustness}A key advantage of the ASVP design lies in its asymmetric interference with the dual path architecture of teacher and student models. During training or inference, the teacher path receives the clean, unperturbed feature map $\mat{X}$, whereas the student path is exposed to its perturbed counterpart $\mat{X}^\prime$. This asymmetric configuration yields three major benefits:

\begin{itemize}
\item[<1>] The teacher model processes clean inputs and maintains full performance integrity;
\item[<2>] The student model only accesses distorted intermediate representations, thereby weakening its ability to imitate informative features;
\item[<3>] The overall structural stability of the training and inference pipeline is preserved, preventing issues such as gradient explosion or representational degradation.
\end{itemize}

In particular, because the teacher never encounters corrupted data, ASVP enables strong defensive interference against KD without retraining or impacting the teacher’s performance, ensuring a robust and non-intrusive protection mechanism.
\begin{comment}

\subsubsection{ Cross Task Generalizability}

The proposed spectral perturbation mechanism is not confined to super resolution or image restoration tasks. It can be extended to a broad range of vision applications where KD is employed, especially in scenarios where the student model attempts to replicate the intermediate representations of the teacher network. Representative examples include object detection and semantic segmentation, which frequently utilize feature based distillation. In such cases, ASVP can be integrated into the backbone of the teacher network to safeguard critical feature information.
\end{comment}
The core requirement for applicability is that the student relies on alignment with the teacher’s internal features. Under this condition, ASVP effectively injects targeted distortions along principal feature directions. Therefore, ASVP serves as a generic and versatile defense strategy for protecting model intellectual property in various computer vision tasks involving intermediate representation distillation.

%%%%%%%%%%%%%%%%%%% tab-1 %%%%%%%%%%%%%%%%%%%%%%%%%%%%%%%%%%%%%
\renewcommand{\arraystretch}{1.3}
\begin{table*}[!htb]
\centering
\caption{\normalsize \textbf{Super-resolution results} on DIV2K with different KD settings. Gray rows show baseline teacher models; blue rows indicate our ASVP method defense.}
\label{tab:SR}
\resizebox{\textwidth}{!}{
\begin{tabular}{l|ccc|ccc|ccc|ccc|ccc}
\specialrule{1pt}{0pt}{2pt}
\multicolumn{1}{c|}{} 
& \multicolumn{3}{c|}{} 
& \multicolumn{12}{c}{\textbf{Student Performance after KD}} \\
\cmidrule(lr){5-16}
\textbf{Teacher Network} 
& \multicolumn{3}{c|}{\textbf{Teacher Performance}} 
& \multicolumn{3}{c|}{\textbf{ResNet9}} 
& \multicolumn{3}{c|}{\textbf{ResNet18}} 
& \multicolumn{3}{c|}{\textbf{SwinIR}}
& \multicolumn{3}{c}{\textbf{AGDN}}\\
\cmidrule(lr){2-4} \cmidrule(lr){5-7} \cmidrule(lr){8-10} \cmidrule(lr){11-13} \cmidrule(lr){14-16}
& PSNR & SSIM & LPIPS & PSNR & SSIM & LPIPS & PSNR & SSIM & LPIPS & PSNR & SSIM & LPIPS & PSNR & SSIM & LPIPS \\
\midrule
Student baseline & -- & -- & -- & 31.63 & 0.801 & 2.100 & 32.07 & 0.831 & 2.067 & 32.37 & 0.851 & 1.885 & 33.09 & 0.897 & 1.681\\

\rowcolor{gray!20}
ResNet18 (normal) & 32.07 & 0.831 & 2.067 & 31.16 & 0.764 & 2.112 & 31.94 & 0.824 & 2.098 & 31.98 & 0.838 & 1.991 & 32.56 & 0.871 & 1.684\\
ResNet18 (noise-L) & 29.89 & 0.641 & 2.921 & 28.73 & 0.342 & 4.289 & 28.81 & 0.422 & 3.251 & 28.77 & 0.503 & 2.917 & 28.85 & 0.451 & 2.699\\
ResNet18 (noise-H) & 28.48 & 0.382 & 4.216 & 28.03 & 0.291 & 4.298 & 28.11 & 0.329 & 4.072 & 28.06 & 0.398 & 3.923 & 28.09 & 0.349 & 4.502\\
ResNet18 (dropC-L) & 32.00 & 0.825 & 2.071 & 31.54 & 0.797 & 2.103 & 31.64 & 0.801 & 2.098 & 31.85 & 0.811 & 1.994 & 32.21 & 0.841 & 1.689\\
ResNet18 (dropC-H) & 30.14 & 0.763 & 2.089 & 31.48 & 0.787 & 2.106 & 31.52 & 0.798 & 2.100 & 31.78 & 0.809 & 1.995 & 32.26 & 0.832 & 1.591\\
ResNet18 (adv-L) & 29.88 & 0.671 & 2.918 & 30.58 & 0.714 & 2.423 & 30.62 & 0.768 & 2.609 & 31.84 & 0.796 & 1.900 & 32.22 & 0.821 & 1.694\\
ResNet18 (adv-H) & 28.02 & 0.094 & 6.634 & 30.51 & 0.707 & 2.529 & 30.93 & 0.772 & 2.511 & 31.92 & 0.758 & 1.997 & 32.25 & 0.832 & 1.695\\

\rowcolor{blue!10}
ResNet18 (ASVP) & 32.07 & 0.831 & 2.067 & 28.01 & 0.106 & 5.337 & 28.04 & 0.142 & 5.312 & 31.05 & 0.706 & 2.362 & 31.56 & 0.721 & 2.544\\

\midrule
\rowcolor{gray!20}
SwinIR (normal) & 32.37 & 0.851 & 1.885 & 31.24 & 0.759 & 2.913 & 32.37 & 0.851 & 2.061 & 32.37 & 0.861 & 1.858 & 32.30 & 0.848 & 1.460\\

\rowcolor{blue!10}
SwinIR (ASVP) & 32.37 & 0.851 & 1.885 & 28.01 & 0.102 & 5.350 & 28.02 & 0.212 & 5.519 & 29.02 & 0.612 & 3.858 & 30.77 & 0.721 & 1.778\\

\midrule
\rowcolor{gray!20}
X-Restormer (normal) & 33.01 & 0.892 & 1.054 & 30.26 & 0.634 & 2.194 & 32.41 & 0.842 & 2.061 & 32.02 & 0.822 & 1.773 & 33.22 & 0.903 & 1.548\\

\rowcolor{blue!10}
X-Restormer (ASVP) & 33.01 & 0.892 & 1.054 & 28.39 & 0.184 & 5.312 & 28.42 & 0.215 & 5.284 & 28.40 & 0.622 & 3.773 & 29.89 & 0.703 & 2.677\\

\midrule
\rowcolor{gray!20}
HAIR (normal) & 32.88 & 0.879 & 1.059 & 29.14 & 0.602 & 2.211 & 30.58 & 0.672 & 2.192 & 31.33 & 0.782 & 1.933 & 32.82 & 0.889 & 1.550\\

\rowcolor{blue!10}
HAIR (ASVP) & 32.88 & 0.879 & 1.059 & 28.15 & 0.128 & 4.245 & 28.44 & 0.251 & 4.183 & 28.35 & 0.634 & 2.933 & 30.54 & 0.753 & 2.374\\

\specialrule{1pt}{0pt}{0pt}
\end{tabular}
}
\end{table*}

%%%%%%%%%%%%%%%%%%% tab-1 %%%%%%%%%%%%%%%%%%%%%%%%%%%%%%%%%%%%%
%%%%%%%%%%%%%%%%%%% tab-2 %%%%%%%%%%%%%%%%%%%%%%%%%%%%%%%%%%%%%
\renewcommand{\arraystretch}{1.3}
\begin{table*}[!htb]
\centering
\caption{\normalsize \textbf{Low light enhancement results} on LOLv1 with different KD  settings. Gray rows show baseline teacher models; blue rows indicate our defense method.}
\label{tab:LLM}
\resizebox{\textwidth}{!}{
\begin{tabular}{l|ccc|ccc|ccc|ccc|ccc}
\specialrule{1pt}{0pt}{2pt}
\multicolumn{1}{c|}{} 
& \multicolumn{3}{c|}{} 
& \multicolumn{12}{c}{\textbf{Student Performance after KD}} \\
\cmidrule(lr){5-16}
\textbf{Teacher Network} 
& \multicolumn{3}{c|}{\textbf{Teacher Performance}} 
& \multicolumn{3}{c|}{\textbf{ResNet9}} 
& \multicolumn{3}{c|}{\textbf{ResNet18}} 
& \multicolumn{3}{c|}{\textbf{SwinIR}} 
& \multicolumn{3}{c}{\textbf{AGDN}} \\
\cmidrule(lr){2-4} \cmidrule(lr){5-7} \cmidrule(lr){8-10} \cmidrule(lr){11-13} \cmidrule(lr){14-16}
& PSNR & SSIM & LPIPS & PSNR & SSIM & LPIPS & PSNR & SSIM & LPIPS & PSNR & SSIM & LPIPS & PSNR & SSIM & LPIPS \\
\midrule
Student baseline & -- & -- & -- & 28.19 & 0.582 & 2.100 & 28.23 & 0.644 & 2.010 & 28.03 & 0.674 & 1.965 & 29.01 & 0.721 & 1.802 \\

\rowcolor{gray!20}
ResNet18 (normal) & 28.23 & 0.644 & 2.010 & 28.24 & 0.586 & 2.093 & 28.25 & 0.676 & 2.015 & 28.05 & 0.634 & 1.985 & 28.99 & 0.713 & 1.864\\
ResNet18 (noise-L) & 28.12 & 0.479 & 2.813 & 28.03 & 0.545 & 2.594 & 28.21 & 0.664 & 2.316 & 27.96 & 0.655 & 2.305 & 28.63 & 0.702 & 1.849 \\
ResNet18 (noise-H) & 27.94 & 0.203 & 3.401 & 27.91 & 0.041 & 6.542 & 27.92 & 0.062 & 6.508 & 27.83 & 0.612 & 2.087 & 28.01 & 0.687 & 1.979 \\
ResNet18 (dropC-L) & 28.09 & 0.599 & 2.031 & 28.23 & 0.578 & 2.104 & 28.33 & 0.676 & 2.017 & 28.02 & 0.621 & 1.992 & 28.87 & 0.697 & 1.840 \\
ResNet18 (dropC-H) & 27.99 & 0.506 & 2.093 & 28.16 & 0.566 & 2.145 & 28.24 & 0.665 & 2.036 & 27.96 & 0.593 & 2.011 & 28.81 & 0.672 & 1.882 \\
ResNet18 (adv-L) & 28.18 & 0.617 & 2.014 & 28.22 & 0.611 & 2.055 & 28.24 & 0.668 & 2.040 & 28.01 & 0.641 & 1.981 & 28.14 & 0.683 & 1.893 \\
ResNet18 (adv-H) & 27.93 & 0.074 & 6.452 & 28.05 & 0.515 & 2.187 & 28.13 & 0.597 & 2.089 & 27.95 & 0.655 & 2.014 & 27.89 & 0.664 & 1.961 \\
\rowcolor{blue!10}
ResNet18 (ASVP) & 28.23 & 0.644 & 2.010 & 27.89 & 0.187 & 5.422 & 27.91 & 0.234 & 5.364 & 27.93 & 0.579 & 3.210 & 26.36 & 0.602 & 2.983 \\

\midrule
\rowcolor{gray!20}
SwinIR (normal) & 28.03 & 0.674 & 1.965 & 28.05 & 0.491 & 2.173 & 28.05 & 0.357 & 2.332 & 28.05 & 0.665 & 1.909 & 28.32 & 0.658 & 1.864 \\
\rowcolor{blue!10}
SwinIR (ASVP) & 28.03 & 0.674 & 1.965 & 27.89 & 0.178 & 5.414 & 27.87 & 0.254 & 5.339 & 27.96 & 0.563 & 3.877 & 26.88 & 0.646 & 2.738 \\

\midrule
\rowcolor{gray!20}
X-Restormer (normal) & 29.35 & 0.699 & 1.832 & 28.67 & 0.643 & 1.953 & 28.84 & 0.666 & 1.887 & 28.87 & 0.663 & 1.839 & 29.25 & 0.685 & 1.761 \\
\rowcolor{blue!10}
X-Restormer (ASVP) & 29.35 & 0.699 & 1.832 & 26.34 & 0.214 & 5.325 & 26.73 & 0.241 & 5.278 & 27.12 & 0.532 & 2.970 & 26.87 & 0.519 & 2.898 \\

\midrule
\rowcolor{gray!20}
HAIR (normal) & 30.04 & 0.703 & 1.803 & 28.74 & 0.626 & 1.964 & 28.77 & 0.668 & 1.873 & 29.12 & 0.675 & 1.841 & 29.45 & 0.699 & 1.739 \\
\rowcolor{blue!10}
HAIR (ASVP) & 30.04 & 0.703 & 1.803 & 27.18 & 0.178 & 5.936 & 27.64 & 0.235 & 4.889 & 27.95 & 0.531 & 3.345 & 28.53 & 0.527 & 3.178 \\

\specialrule{1pt}{0pt}{0pt}
\end{tabular}
}
\end{table*}
%%%%%%%%%%%%%%%%%%% tab-2 %%%%%%%%%%%%%%%%%%%%%%%%%%%%%%%%%%%%%

%%%%%%%%%%%%%%%%%%% tab-3 %%%%%%%%%%%%%%%%%%%%%%%%%%%%%%%%%%%%%
\renewcommand{\arraystretch}{1.3}
\begin{table*}[!htb]
\centering
\caption{\normalsize \textbf{Underwater enhancement results} on LSUI with different KD settings. Gray rows show baseline teacher models; blue rows indicate our defense method.}
\label{tab:UW}
\resizebox{\textwidth}{!}{
\begin{tabular}{l|ccc|ccc|ccc|ccc | ccc}
\specialrule{1pt}{0pt}{2pt}
\multicolumn{1}{c|}{} 
& \multicolumn{3}{c|}{} 
& \multicolumn{12}{c}{\textbf{Student Performance after KD}} \\
\cmidrule(lr){5-16}
\textbf{Teacher Network} 
& \multicolumn{3}{c|}{\textbf{Teacher Performance}} 
& \multicolumn{3}{c|}{\textbf{ResNet9}} 
& \multicolumn{3}{c|}{\textbf{ResNet18}} 
& \multicolumn{3}{c|}{\textbf{SwinIR}} 
& \multicolumn{3}{c}{\textbf{AGDN}} \\
\cmidrule(lr){2-4} \cmidrule(lr){5-7} \cmidrule(lr){8-10} \cmidrule(lr){11-13} \cmidrule(lr){14-16}
& PSNR & SSIM & LPIPS & PSNR & SSIM & LPIPS & PSNR & SSIM & LPIPS & PSNR & SSIM & LPIPS  & PSNR & SSIM & LPIPS\\
\midrule
Student baseline        & --    & --    & --    & 28.65 & 0.772 & 2.150 & 28.77 & 0.774 & 2.100 & 29.00 & 0.818 & 2.030 & 29.11 & 0.825 & 1.950\\
\rowcolor{gray!20}
ResNet18 (normal)       & 28.77 & 0.774 & 2.100 & 28.63 & 0.768 & 2.150 & 28.81 & 0.787 & 2.125 & 28.84 & 0.772 & 2.080 & 29.04 & 0.812 & 1.950\\
ResNet18 (noise-L) & 28.51 & 0.551 & 2.230 & 28.69 & 0.693 & 2.250 & 28.77 & 0.766 & 2.150 & 28.74 & 0.736 & 2.080 & 28.91 & 0.804 & 1.950\\
ResNet18 (noise-H)   & 27.84 & 0.014 & 5.400 & 27.92 & 0.098 & 4.500 & 27.92 & 0.131 & 4.400 & 28.24 & 0.645 & 2.150 & 28.67 & 0.784 & 2.030\\
ResNet18 (dropC-L)    & 28.74 & 0.743 & 2.160 & 28.56 & 0.744 & 2.200 & 28.79 & 0.771 & 2.125 & 28.70 & 0.701 & 2.090 & 28.89 & 0.800 & 1.970\\
ResNet18 (dropC-H)    & 28.55 & 0.717 & 2.180 & 28.43 & 0.735 & 2.210 & 28.66 & 0.722 & 2.180 & 28.69 & 0.712 & 2.140 & 29.83 & 0.788 & 1.920\\
ResNet18 (adv-L)     & 28.76 & 0.752 & 2.150 & 28.69 & 0.764 & 2.180 & 28.88 & 0.771 & 2.160 & 28.74 & 0.742 & 2.130 & 28.75 & 0.779 & 2.010\\
ResNet18 (adv-H)     & 27.93 & 0.081 & 6.500 & 28.49 & 0.728 & 2.220 & 28.53 & 0.745 & 2.190 & 28.75 & 0.723 & 2.100 & 28.70 & 0.762 & 2.050\\
\rowcolor{blue!10}
ResNet18 (ASVP)          & 28.77 & 0.774 & 2.100 & 27.84 & 0.154 & 5.500 & 27.86 & 0.178 & 5.480 & 28.08 & 0.612 & 3.430 & 28.27 & 0.682 & 2.980\\
\midrule
\rowcolor{gray!20}
SwinIR (normal)         & 29.00 & 0.818 & 2.030 & 28.01 & 0.799 & 2.250 & 28.04 & 0.803 & 2.230 & 28.71 & 0.802 & 2.180 & 29.18 & 0.828 & 2.090\\
\rowcolor{blue!10}
SwinIR (ASVP)            & 29.00 & 0.818 & 2.030 & 27.56 & 0.143 & 5.500 & 27.82 & 0.196 & 5.450 & 28.44 & 0.743 & 3.380 & 28.09 & 0.790 & 3.290\\
\midrule
\rowcolor{gray!20}
X-Restormer (normal)    & 29.54 & 0.834 & 1.920 & 28.28 & 0.812 & 2.200 & 28.84 & 0.844 & 2.190 & 28.87 & 0.849 & 2.140 & 29.25 & 0.832 & 2.050\\
\rowcolor{blue!10}
X-Restormer (ASVP)      & 29.54 & 0.834 & 1.920 & 26.34 & 0.214 & 5.600 & 26.73 & 0.241 & 5.500 & 27.12 & 0.532 & 4.480 & 28.87 & 0.719 & 3.750\\
\midrule
\rowcolor{gray!20}
HAIR (normal)         & 29.97 & 0.875 & 1.890 & 28.65 & 0.778 & 2.250 & 28.89 & 0.823 & 2.220 & 29.09 & 0.875 & 2.160 & 29.35 & 0.877 & 2.050\\
\rowcolor{blue!10}
HAIR (ASVP)           & 29.97 & 0.875 & 1.890 & 27.10 & 0.175 & 4.800 & 27.47 & 0.234 & 4.750 & 27.55 & 0.737 & 3.700 & 27.13 & 0.753 & 3.650\\
\specialrule{1pt}{0pt}{0pt}
\end{tabular}
}
\end{table*}

%%%%%%%%%%%%%%%%%%% tab-3 %%%%%%%%%%%%%%%%%%%%%%%%%%%%%%%%%%%%%

%%%%%%%%%%%%%%%%%%% tab-4 %%%%%%%%%%%%%%%%%%%%%%%%%%%%%%%%%%%%%
\renewcommand{\arraystretch}{1.3}
\begin{table*}[!htb]
\centering
\caption{\normalsize \textbf{Dehazing results} on SOTS with different KD  settings. Gray rows show baseline teacher models; blue rows indicate our  defense method.}
\label{tab:Dehaze}
\resizebox{\textwidth}{!}{
\begin{tabular}{l|ccc|ccc|ccc|ccc | ccc}
\specialrule{1pt}{0pt}{2pt}
\multicolumn{1}{c|}{} 
& \multicolumn{3}{c|}{} 
& \multicolumn{12}{c}{\textbf{Student Performance after KD}} \\
\cmidrule(lr){5-16}
\textbf{Teacher Network} 
& \multicolumn{3}{c|}{\textbf{Teacher Performance}} 
& \multicolumn{3}{c|}{\textbf{ResNet9}} 
& \multicolumn{3}{c|}{\textbf{ResNet18}} 
& \multicolumn{3}{c|}{\textbf{SwinIR}} 
& \multicolumn{3}{c}{\textbf{AGDN}} \\
\cmidrule(lr){2-4} \cmidrule(lr){5-7} \cmidrule(lr){8-10} \cmidrule(lr){11-13} \cmidrule(lr){14-16}
& PSNR & SSIM & LPIPS & PSNR & SSIM & LPIPS & PSNR & SSIM & LPIPS & PSNR & SSIM & LPIPS  & PSNR & SSIM & LPIPS\\
\midrule
Student baseline        & --    & --    & --    & 28.57 & 0.818 & 2.080 & 28.77 & 0.774 & 2.120 & 29.35 & 0.851 & 2.030 & 29.62 & 0.881 & 1.980\\
\rowcolor{gray!20}
ResNet18 (normal)       & 28.77 & 0.774 & 2.120 & 28.66 & 0.817 & 2.090 & 28.78 & 0.790 & 2.110 & 29.26 & 0.811 & 2.050 & 29.42 & 0.839 & 2.020\\
ResNet18 (noise-L) & 28.41 & 0.411 & 3.200 & 27.87 & 0.181 & 6.250 & 28.00 & 0.291 & 6.130 & 29.03 & 0.713 & 1.960 & 29.37 & 0.833 & 1.890\\
ResNet18 (noise-H)   & 28.37 & 0.283 & 5.300 & 27.93 & 0.135 & 6.330 & 28.01 & 0.266 & 6.280 & 28.78 & 0.706 & 2.020 & 29.12 & 0.826 & 1.930\\
ResNet18 (dropC-L)    & 28.10 & 0.729 & 2.160 & 28.58 & 0.794 & 2.210 & 28.60 & 0.826 & 2.200 & 29.12 & 0.784 & 2.100 & 29.40 & 0.831 & 2.070\\
ResNet18 (dropC-H)    & 28.55 & 0.717 & 2.180 & 28.43 & 0.735 & 2.190 & 28.66 & 0.722 & 2.160 & 28.69 & 0.712 & 2.110 & 29.83 & 0.788 & 2.050\\
ResNet18 (adv-L)     & 28.76 & 0.752 & 2.160 & 28.69 & 0.764 & 2.190 & 28.88 & 0.771 & 2.180 & 29.12 & 0.772 & 2.140 & 29.25 & 0.830 & 2.070\\
ResNet18 (adv-H)     & 27.86 & 0.174 & 5.400 & 28.08 & 0.794 & 2.230 & 28.18 & 0.800 & 2.210 & 29.05 & 0.769 & 2.080 & 29.21 & 0.822 & 2.050\\
\rowcolor{blue!10}
ResNet18 (ASVP)          & 28.77 & 0.774 & 2.120 & 27.87 & 0.188 & 5.400 & 27.90 & 0.243 & 5.380 & 28.54 & 0.645 & 3.280 & 28.04 & 0.651 & 3.200\\
\midrule
\rowcolor{gray!20}
SwinIR (normal)         & 29.35 & 0.851 & 2.030 & 28.11 & 0.799 & 2.160 & 28.33 & 0.803 & 2.150 & 29.25 & 0.826 & 2.100 & 29.45 & 0.863 & 2.050\\
\rowcolor{blue!10}
SwinIR (ASVP)            & 29.35 & 0.851 & 2.030 & 27.85 & 0.283 & 4.450 & 27.91 & 0.212 & 5.420 & 28.14 & 0.756 & 3.340 & 28.55 & 0.768 & 3.280\\
\midrule
\rowcolor{gray!20}
X-Restormer (normal)    & 31.68 & 0.873 & 1.800 & 29.25 & 0.819 & 2.070 & 29.64 & 0.824 & 2.050 & 30.05 & 0.849 & 2.020 & 30.25 & 0.856 & 1.980\\
\rowcolor{blue!10}
X-Restormer (ASVP)      & 31.68 & 0.873 & 1.800 & 27.39 & 0.216 & 5.500 & 27.45 & 0.244 & 5.470 & 27.87 & 0.639 & 3.910 & 27.47 & 0.719 & 3.480\\
\midrule
\rowcolor{gray!20}
HAIR (normal)         & 31.05 & 0.866 & 1.750 & 28.95 & 0.799 & 2.050 & 29.11 & 0.812 & 2.030 & 29.34 & 0.846 & 2.000 & 30.24 & 0.858 & 1.950\\
\rowcolor{blue!10}
HAIR (ASVP)           & 31.05 & 0.866 & 1.750 & 27.10 & 0.176 & 4.850 & 27.44 & 0.204 & 4.820 & 28.25 & 0.673 & 3.770 & 27.13 & 0.755 & 3.440\\

\specialrule{1pt}{0pt}{0pt}
\end{tabular}
}
\end{table*}
%%%%%%%%%%%%%%%%%%% tab-4 %%%%%%%%%%%%%%%%%%%%%%%%%%%%%%%%%%%%%

%%%%%%%%%%%%%%%%%%% tab-5 %%%%%%%%%%%%%%%%%%%%%%%%%%%%%%%%%%%%%
\renewcommand{\arraystretch}{1.3}
\begin{table}[!htb]
\centering
\caption{\normalsize \textbf{Deraining results} on Rain100 with different KD  settings. Gray rows show baseline teacher models; blue rows indicate our  defense method.}
\label{tab:Derain}
\resizebox{\textwidth}{!}{
\begin{tabular}{l|ccc|ccc|ccc|ccc | ccc}
\specialrule{1pt}{0pt}{2pt}
\multicolumn{1}{c|}{} 
& \multicolumn{3}{c|}{} 
& \multicolumn{12}{c}{\textbf{Student Performance after KD}} \\
\cmidrule(lr){5-16}
\textbf{Teacher Network} 
& \multicolumn{3}{c|}{\textbf{Teacher Performance}} 
& \multicolumn{3}{c|}{\textbf{ResNet9}} 
& \multicolumn{3}{c|}{\textbf{ResNet18}} 
& \multicolumn{3}{c|}{\textbf{SwinIR}} 
& \multicolumn{3}{c}{\textbf{AGDN}} \\
\cmidrule(lr){2-4} \cmidrule(lr){5-7} \cmidrule(lr){8-10} \cmidrule(lr){11-13}  \cmidrule(lr){14-16}
& PSNR & SSIM & LPIPS & PSNR & SSIM & LPIPS & PSNR & SSIM & LPIPS & PSNR & SSIM & LPIPS  & PSNR & SSIM & LPIPS\\
\midrule
Student baseline        & --    & --    & --    & 28.96 & 0.634 & 2.100 & 29.12 & 0.642 & 2.050 & 29.35 & 0.851 & 1.710 & 29.34 & 0.704 & 2.010 \\
\rowcolor{gray!20}
ResNet18 (normal)       & 29.12 & 0.642 & 2.050 & 29.02 & 0.622 & 2.100 & 29.04 & 0.638 & 2.100 & 29.26 & 0.811 & 2.030 & 29.42 & 0.839 & 2.010 \\
ResNet18 (noise-L) & 28.62 & 0.512 & 2.910 & 28.33 & 0.403 & 3.280 & 28.39 & 0.431 & 3.130 & 28.71 & 0.604 & 2.050 & 28.88 & 0.676 & 1.990 \\
ResNet18 (noise-H)   & 28.38 & 0.377 & 4.300 & 27.98 & 0.067 & 6.750 & 28.00 & 0.132 & 6.300 & 28.68 & 0.582 & 2.100 & 28.69 & 0.664 & 2.010 \\
ResNet18 (dropC-L)    & 28.92 & 0.652 & 2.150 & 28.97 & 0.622 & 2.170 & 28.99 & 0.633 & 2.140 & 28.75 & 0.624 & 2.080 & 29.11 & 0.684 & 2.050 \\
ResNet18 (dropC-H)    & 28.89 & 0.543 & 2.170 & 28.86 & 0.610 & 2.190 & 28.95 & 0.619 & 2.160 & 28.72 & 0.612 & 2.120 & 29.02 & 0.682 & 2.010 \\
ResNet18 (adv-L)     & 28.39 & 0.555 & 2.200 & 28.57 & 0.604 & 2.220 & 28.99 & 0.617 & 2.180 & 28.65 & 0.638 & 2.140 & 28.78 & 0.654 & 2.030 \\
ResNet18 (adv-H)     & 27.97 & 0.017 & 6.350 & 28.65 & 0.548 & 2.270 & 28.72 & 0.558 & 2.240 & 28.54 & 0.599 & 2.100 & 28.62 & 0.644 & 2.010 \\
\rowcolor{blue!10}
ResNet18 (ASVP)          & 29.12 & 0.642 & 2.050 & 27.96 & 0.115 & 5.400 & 27.98 & 0.110 & 5.350 & 28.14 & 0.553 & 3.720 & 27.59 & 0.604 & 3.150 \\
\midrule
\rowcolor{gray!20}
SwinIR (normal)         & 29.35 & 0.851 & 1.710 & 28.11 & 0.799 & 2.250 & 28.33 & 0.803 & 2.220 & 29.25 & 0.826 & 2.150 & 29.45 & 0.863 & 2.100 \\
\rowcolor{blue!10}
SwinIR (ASVP)            & 29.35 & 0.851 & 1.710 & 27.85 & 0.283 & 5.300 & 27.91 & 0.245 & 5.270 & 28.02 & 0.512 & 3.910 & 27.78 & 0.629 & 3.143 \\
\midrule
\rowcolor{gray!20}
X-Restormer (normal)    & 30.21 & 0.773 & 1.900 & 29.00 & 0.698 & 2.150 & 29.23 & 0.723 & 2.140 & 29.73 & 0.745 & 2.120 & 30.25 & 0.856 & 2.100 \\
\rowcolor{blue!10}
X-Restormer (ASVP)      & 30.21 & 0.773 & 1.900 & 28.32 & 0.213 & 5.400 & 28.45 & 0.276 & 5.350 & 28.82 & 0.632 & 3.300 & 28.17 & 0.719 & 2.950 \\
\midrule
\rowcolor{gray!20}
HAIR (normal)         & 29.68 & 0.713 & 1.800 & 28.55 & 0.644 & 2.120 & 28.68 & 0.678 & 2.100 & 28.79 & 0.690 & 2.080 & 29.24 & 0.755 & 2.050 \\
\rowcolor{blue!10}
HAIR (ASVP)           & 29.68 & 0.713 & 1.800 & 27.11 & 0.172 & 5.950 & 27.47 & 0.254 & 4.910 & 27.85 & 0.612 & 3.560 & 27.13 & 0.708 & 2.800 \\

\specialrule{1pt}{0pt}{0pt}
\end{tabular}
}
\end{table}
%%%%%%%%%%%%%%%%%%% tab-5 %%%%%%%%%%%%%%%%%%%%%%%%%%%%%%%%%%%%%

\section{EXPERIMENTS AND ANALYSIS}
\subsection{Experimental Setup}
We evaluate our ASVP defense across five representative image restoration tasks: single image super resolution, low light enhancement, underwater image enhancement, dehazing and deraining. The teacher models are two high capacity networks (ResNet-18\cite{qin2020deep_Resnet} and SwinIR\cite{liang2021_swinir}), and the student models(ResNet-9 or ResNet-18 and SwinIR) trained to mimic the teachers. Each task uses a standard dataset (e.g., DIV2K for super-resolution, LOLv1 for low light, LSUI for underwater, SOTS for dehazing, Rain100 for deraining) and identical training protocols. We measure restoration quality using PSNR (peak signal-to-noise ratio) \cite{huynh2008scope_PSNR} and SSIM (structural similarity index)\cite{wang2004image_SSIM}, which quantify pixel wise fidelity and perceptual similarity, respectively. Higher PSNR or SSIM indicates better reconstruction: PSNR is typically in decibels (dB) and higher values mean smaller reconstruction error; SSIM ranges from 0 to 1, with values closer to 1 indicating images more similar in structure to the reference.

We compare three scenarios for each teacher student pair: (1) Baseline KD (no defense): the student is distilled from the teacher’s outputs as usual; (2) Other defenses: we consider naive feature perturbations (noise, channel drop, adversarial perturbations), where “L” and “H” denote low and high perturbation intensity levels, respectively; (3) ASVP defense: our adaptive singular value perturbation module is applied to the teacher during inference. All student models are then trained on the  teacher outputs. Both teacher and student networks are trained using L1 loss, with student training further guided by a combination of L1 and perceptual losses. Optimization is performed using Adam($\beta_1=0.9$,$\beta_2=0.999$), with a learning rate of 1e-4, batch size of 16 (ResNet) or 8 (SwinIR), and 350 total epochs.

%\cref{tab:SR,tab:LLM,tab:UW,tab:Dehaze,tab:Derain}  summarize the PSNR and SSIM of both teacher and student models for each task and scenario. Gray rows denote the undefended teachers (baseline), while blue rows denote results with our ASVP defense.

%%%%%%%%%%%%%%%%%%% fig-1 %%%%%%%%%%%%%%%%%%%%%%%%%%%%%%%%%%%%%
\setlength{\tabcolsep}{1pt}
\def \imgl {0.14\linewidth}
\def \imgs {0.15\linewidth}
\begin{figure*}[!htb]
\small
\begin{center}
\begin{tabular}{ccccccc}
\rotatebox{90}{\textbf{Teacher}} &
\includegraphics[width=\imgs]{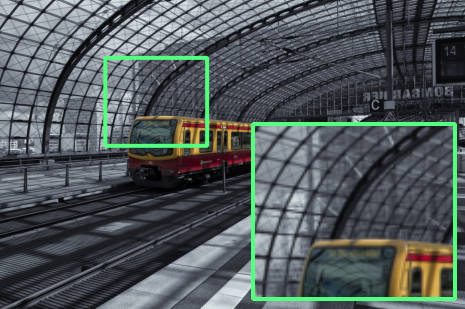} &
\includegraphics[width=\imgs]{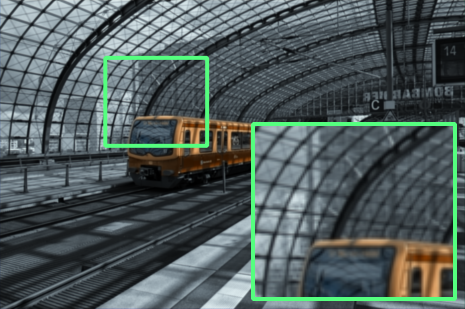} &
\includegraphics[width=\imgs]{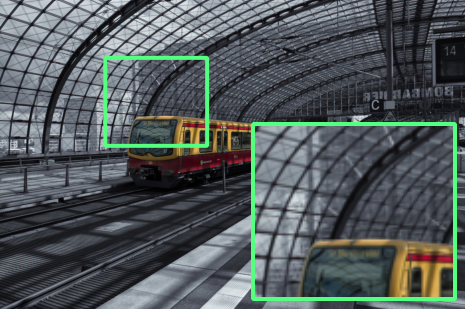} &
\includegraphics[width=\imgs]{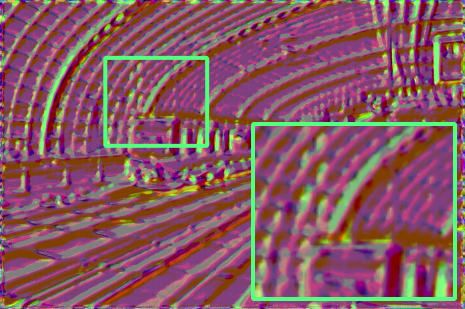} &
\includegraphics[width=\imgs]{fig/SR/0202/gt-drop-svd-swin_processed.png} & 
\includegraphics[width=\imgs]{fig/SR/0202/gt-drop-svd-swin_processed.png}  \\
\rotatebox{90}{\textbf{KD}} &
\includegraphics[width=\imgs]{fig/SR/0202/kd-gt-kd-drop_processed.png} &
\includegraphics[width=\imgs]{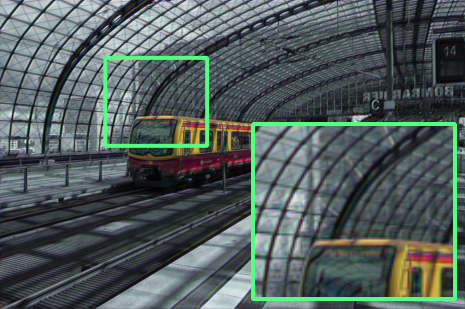} &
\includegraphics[width=\imgs]{fig/SR/0202/kd-gt-kd-drop_processed.png} &
\includegraphics[width=\imgs]{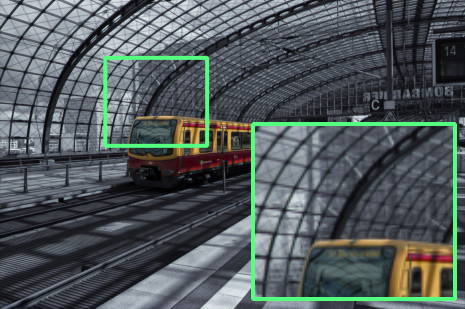} &
\includegraphics[width=\imgs]{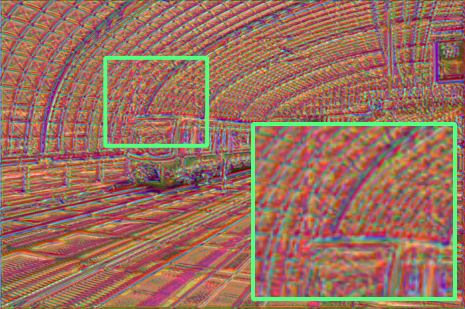} &
\includegraphics[width=\imgs]{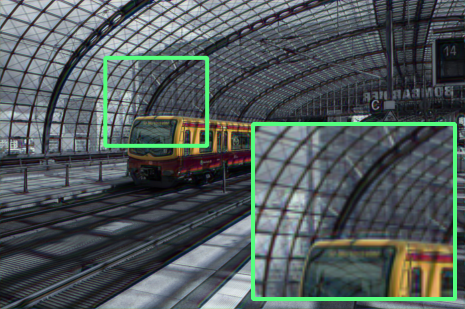} \\

\rotatebox{90}{\textbf{Teacher}} &
\includegraphics[width=\imgs]{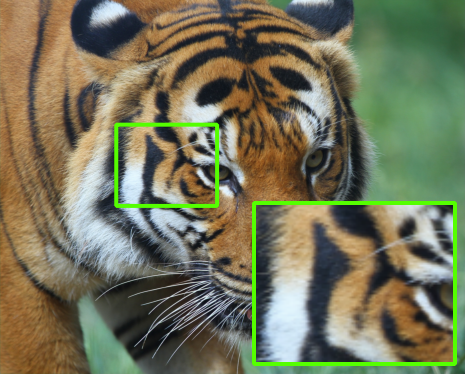} &
\includegraphics[width=\imgs]{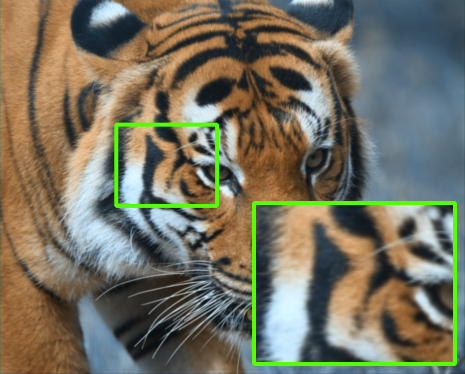} &
\includegraphics[width=\imgs]{fig/SR/0010/gt_processed.png} &
\includegraphics[width=\imgs]{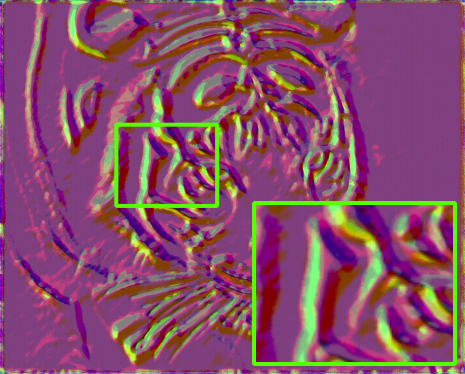} &
\includegraphics[width=\imgs]{fig/SR/0010/gt_processed.png} & 
\includegraphics[width=\imgs]{fig/SR/0010/gt_processed.png}  \\
\rotatebox{90}{\textbf{KD}} &
\includegraphics[width=\imgs]{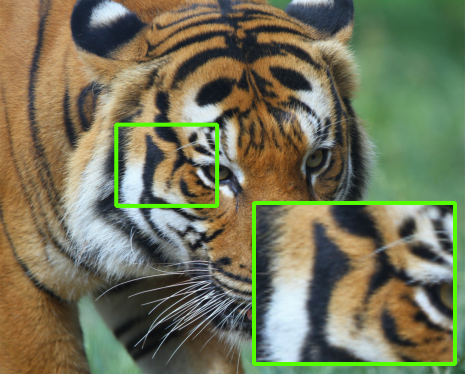} &
\includegraphics[width=\imgs]{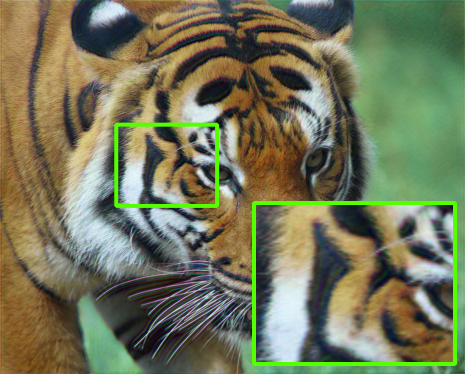} &
\includegraphics[width=\imgs]{fig/SR/0010/kd-gt_processed.png} &
\includegraphics[width=\imgs]{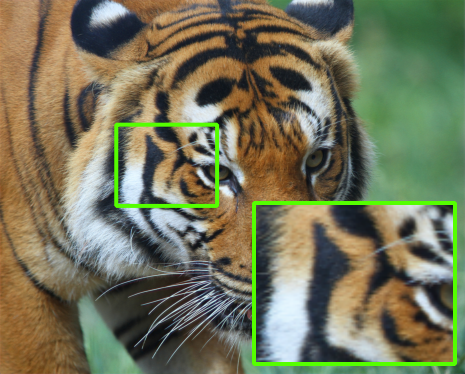} &
\includegraphics[width=\imgs]{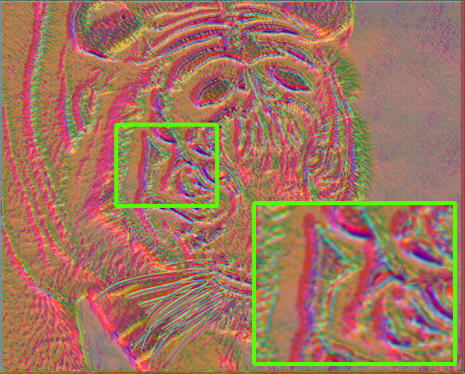} &
\includegraphics[width=\imgs]{fig/SR/0010/noise08_processed.png} \\
& normal & noise & dropC & adv & ResNet-ASVp & Swin-ASVP \\
\end{tabular}
\end{center}\vspace{-2mm}
\caption{Visual results of super-resolution. Our method degrades student outputs while maintaining high fidelity reconstruction in the teacher.}

\vspace{-2mm}
\label{fig:SR}
\end{figure*}
%%%%%%%%%%%%%%%%%%% fig-1 %%%%%%%%%%%%%%%%%%%%%%%%%%%%%%%%%%%%%

%%%%%%%%%%%%%%%%%%% fig-2 %%%%%%%%%%%%%%%%%%%%%%%%%%%%%%%%%%%%%
\setlength{\tabcolsep}{1pt}
\def \imgl {0.14\linewidth}
\def \imgs {0.145\linewidth}
\begin{figure*}[!htb]
\small
\begin{center}
\begin{tabular}{ccccccc}
\rotatebox{90}{\textbf{Teacher}} &
\includegraphics[width=\imgs]{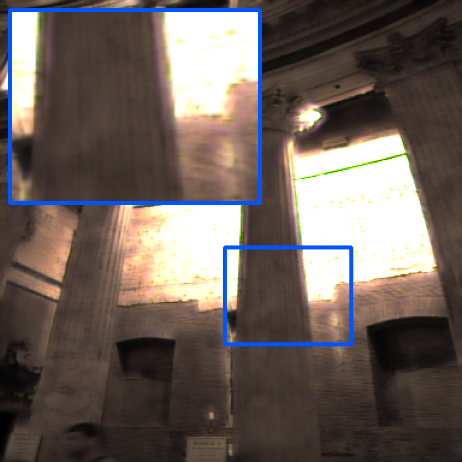} &
\includegraphics[width=\imgs]{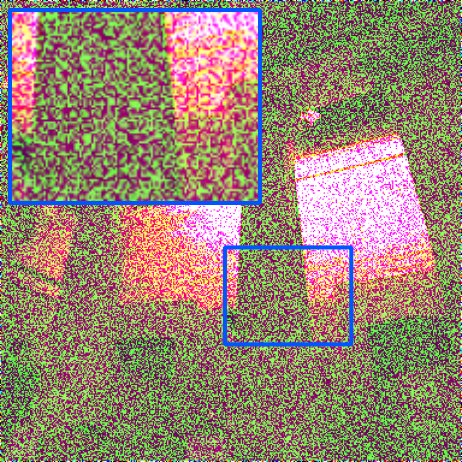} &
\includegraphics[width=\imgs]{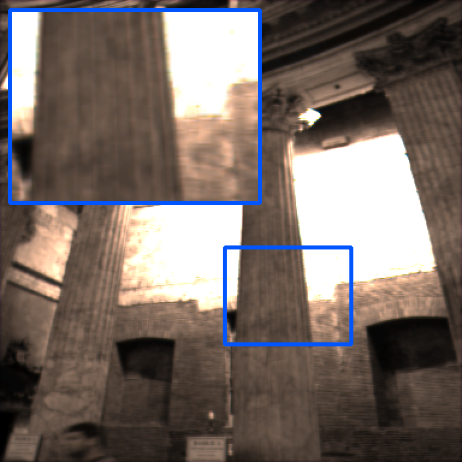} &
\includegraphics[width=\imgs]{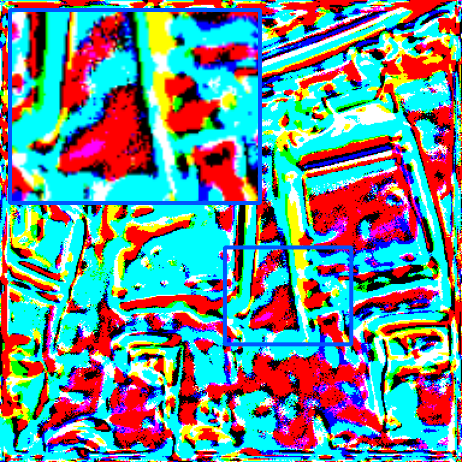} &
\includegraphics[width=\imgs]{fig/LLM/1/gt-svd_processed.png} & 
\includegraphics[width=\imgs]{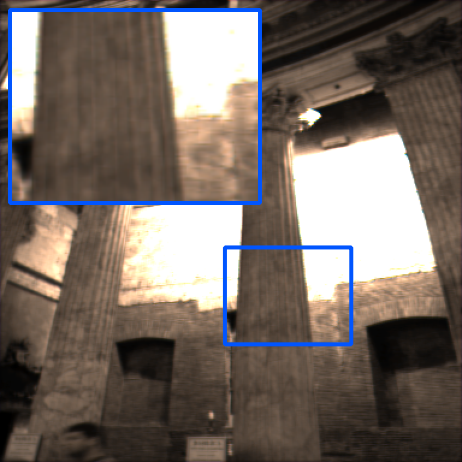}  \\
\rotatebox{90}{\textbf{KD}} &
\includegraphics[width=\imgs]{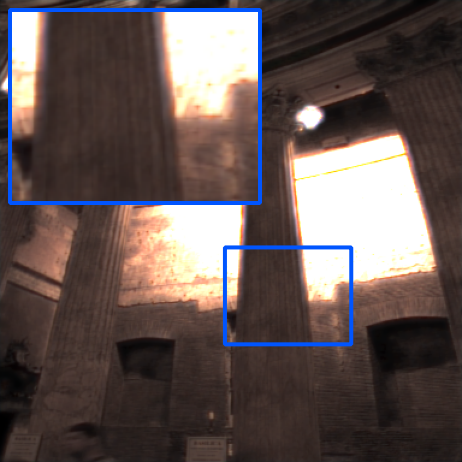} &
\includegraphics[width=\imgs]{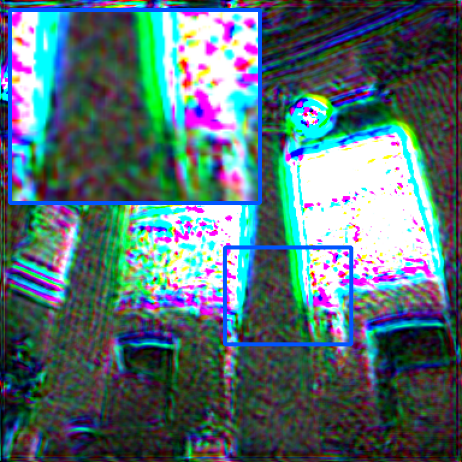} &
\includegraphics[width=\imgs]{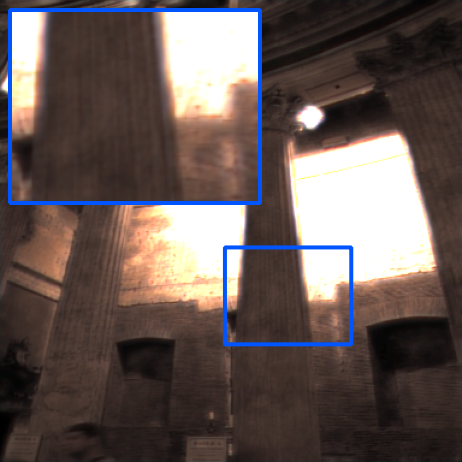} &
\includegraphics[width=\imgs]{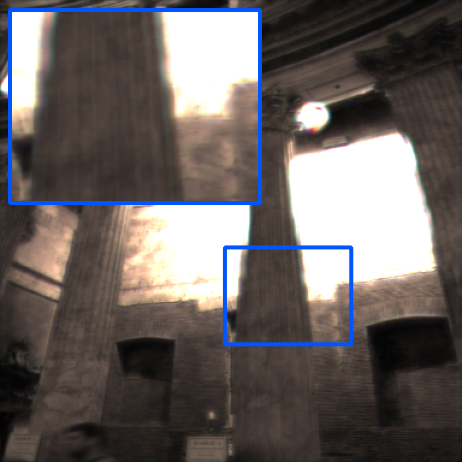} &
\includegraphics[width=\imgs]{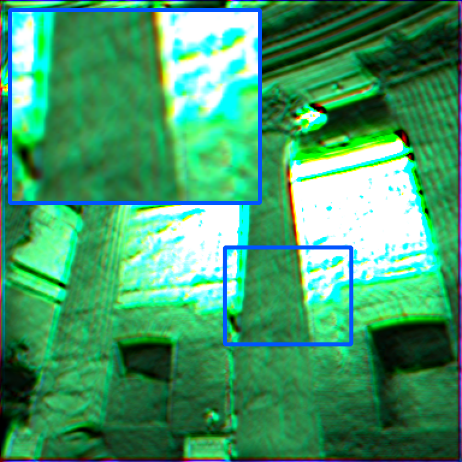} &
\includegraphics[width=\imgs]{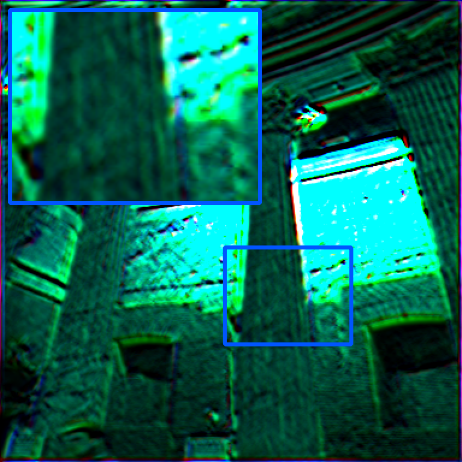} \\

\rotatebox{90}{\textbf{Teacher}} &
\includegraphics[width=\imgs]{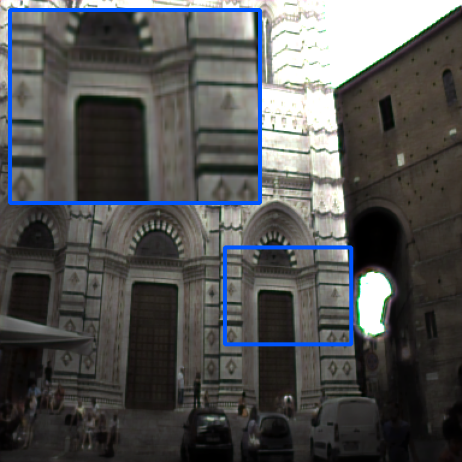} &
\includegraphics[width=\imgs]{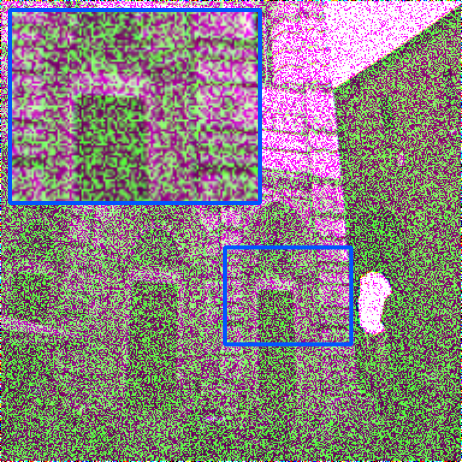} &
\includegraphics[width=\imgs]{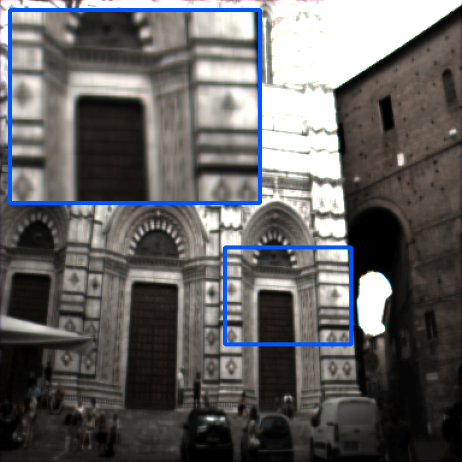} &
\includegraphics[width=\imgs]{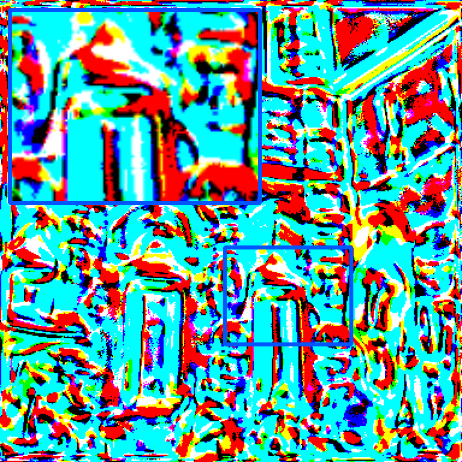} &
\includegraphics[width=\imgs]{fig/LLM/4/gt-svd_processed.png} & 
\includegraphics[width=\imgs]{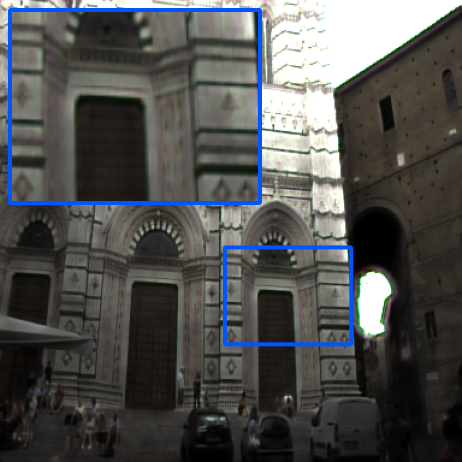}  \\
\rotatebox{90}{\textbf{KD}} &
\includegraphics[width=\imgs]{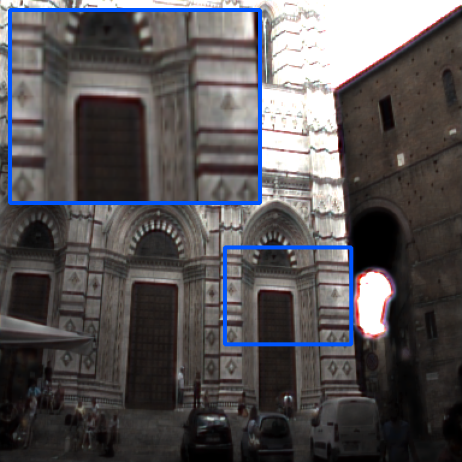} &
\includegraphics[width=\imgs]{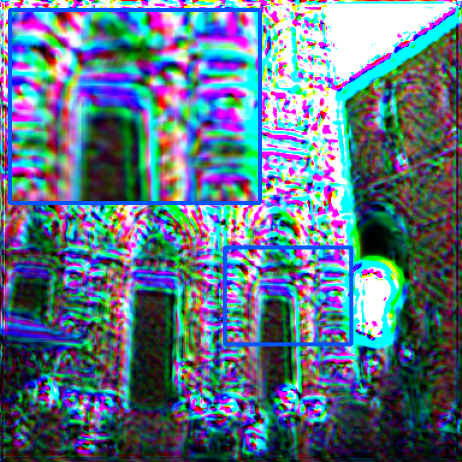} &
\includegraphics[width=\imgs]{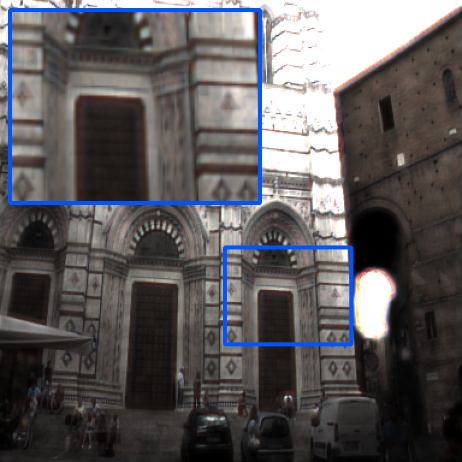} &
\includegraphics[width=\imgs]{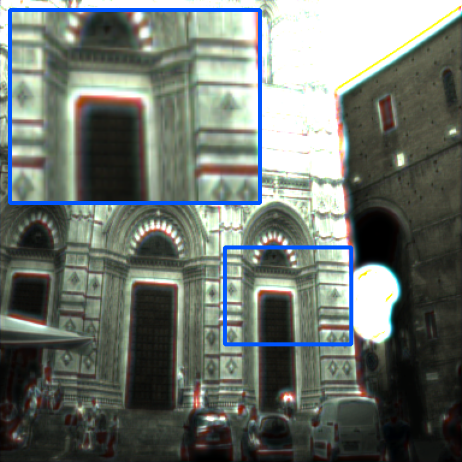} &
\includegraphics[width=\imgs]{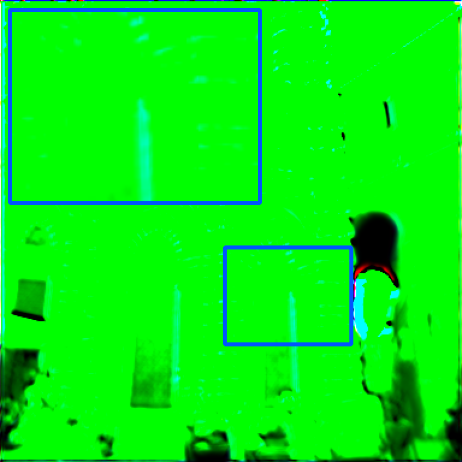} &
\includegraphics[width=\imgs]{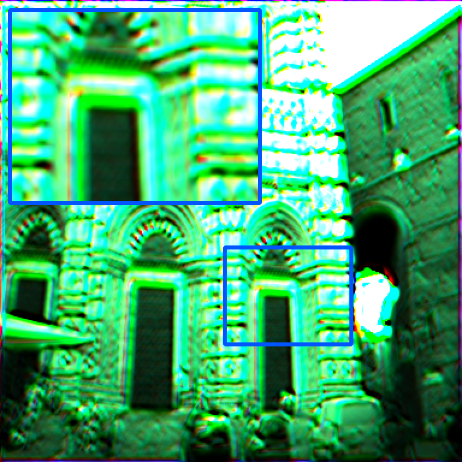} \\
& normal & noise & dropC & adv & ResNet-ASVP & Swin-ASVP \\
\end{tabular}
\end{center}\vspace{-2mm}
\caption{Low-light enhancement results. Our method suppresses student outputs while preserving teacher brightness and detail.}

\vspace{-2mm}
\label{fig:LLM}
\end{figure*}
%%%%%%%%%%%%%%%%%%% fig-2 %%%%%%%%%%%%%%%%%%%%%%%%%%%%%%%%%%%%%

%%%%%%%%%%%%%%%%%%% fig-3 %%%%%%%%%%%%%%%%%%%%%%%%%%%%%%%%%%%%%
\setlength{\tabcolsep}{1pt}
\def \imgl {0.14\linewidth}
\def \imgs {0.145\linewidth}
\begin{figure*}[!htb]
\small
\begin{center}
\begin{tabular}{ccccccc}
\rotatebox{90}{\textbf{Teacher}} &
\includegraphics[width=\imgs]{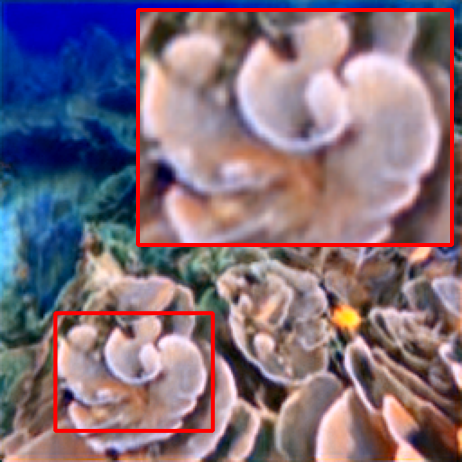} &
\includegraphics[width=\imgs]{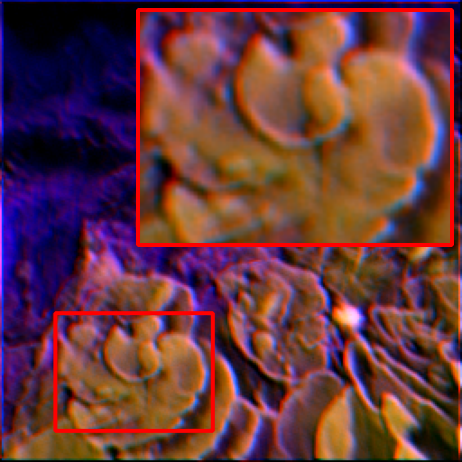} &
\includegraphics[width=\imgs]{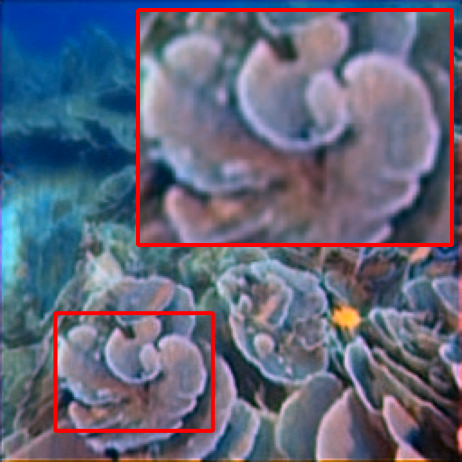} &
\includegraphics[width=\imgs]{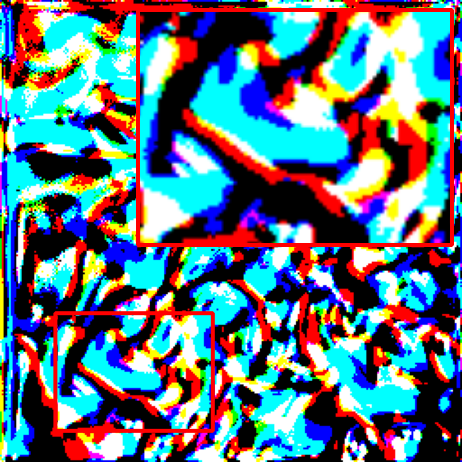} &
\includegraphics[width=\imgs]{fig/UW/2012/gt.-svd_processed.png} & 
\includegraphics[width=\imgs]{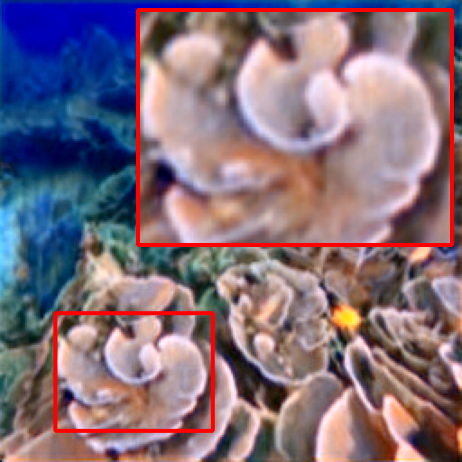}  \\
\rotatebox{90}{\textbf{KD}} &
\includegraphics[width=\imgs]{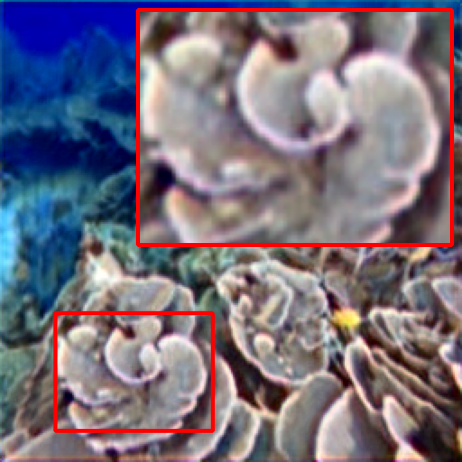} &
\includegraphics[width=\imgs]{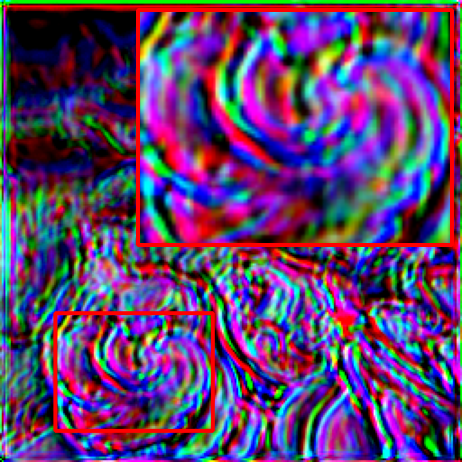} &
\includegraphics[width=\imgs]{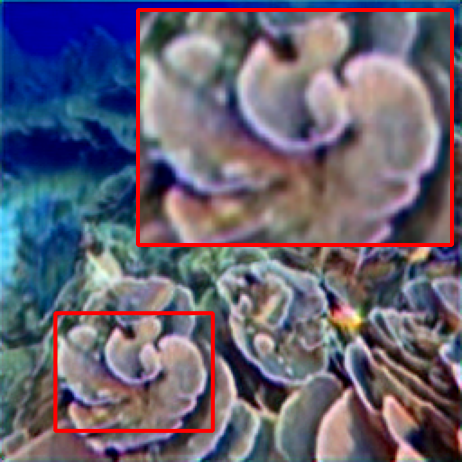} &
\includegraphics[width=\imgs]{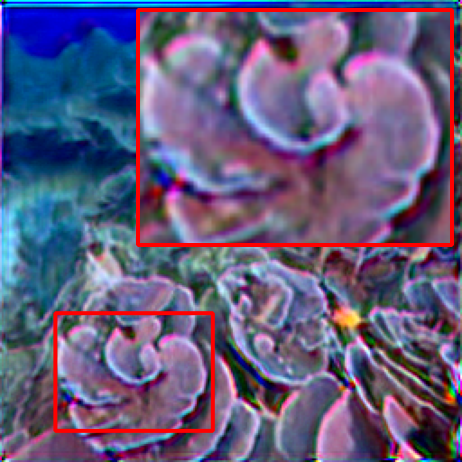} &
\includegraphics[width=\imgs]{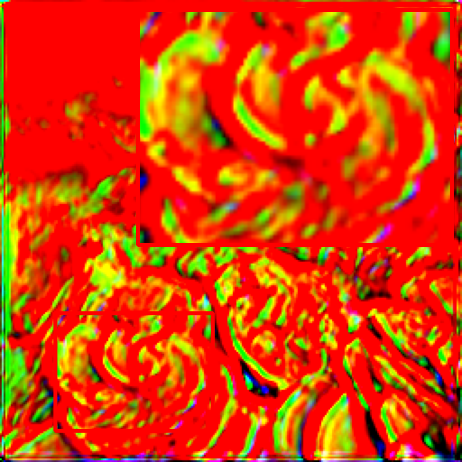} &
\includegraphics[width=\imgs]{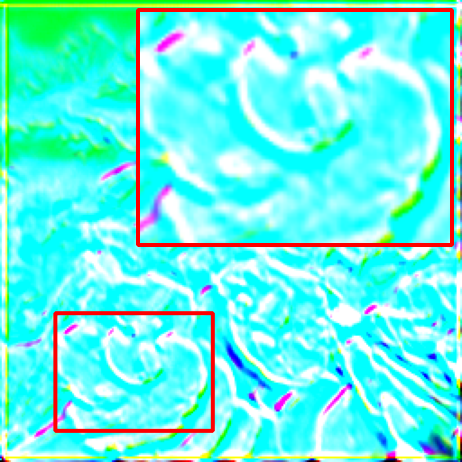} \\

\rotatebox{90}{\textbf{Teacher}} &
\includegraphics[width=\imgs]{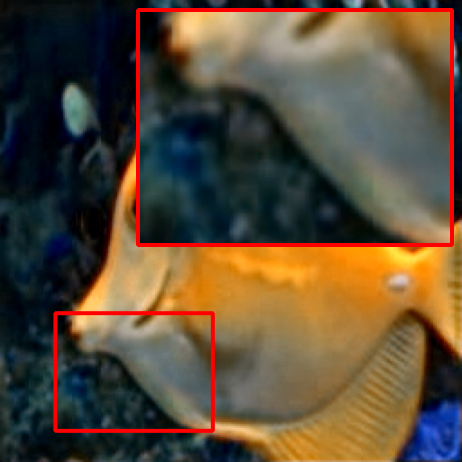} &
\includegraphics[width=\imgs]{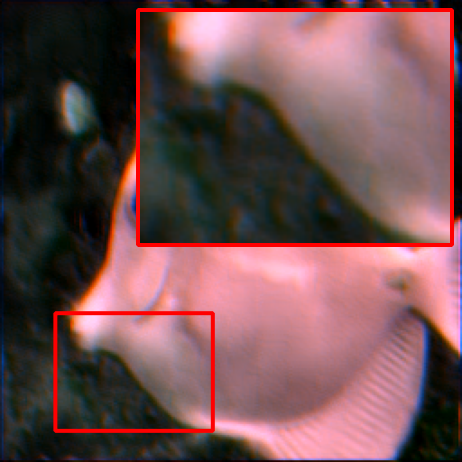} &
\includegraphics[width=\imgs]{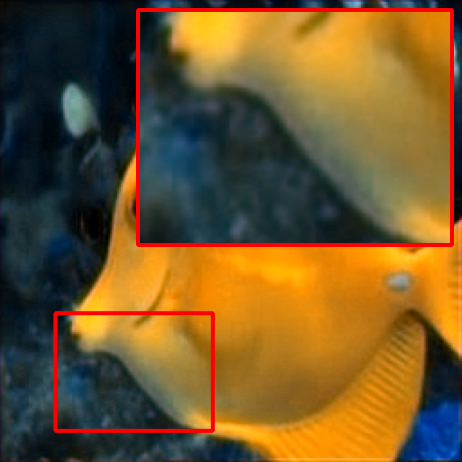} &
\includegraphics[width=\imgs]{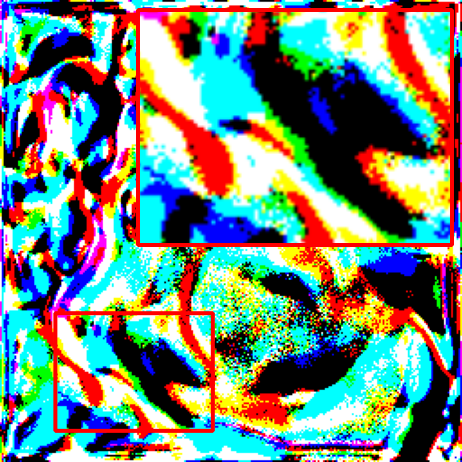} &
\includegraphics[width=\imgs]{fig/UW/2071/gt-svd_processed.png} & 
\includegraphics[width=\imgs]{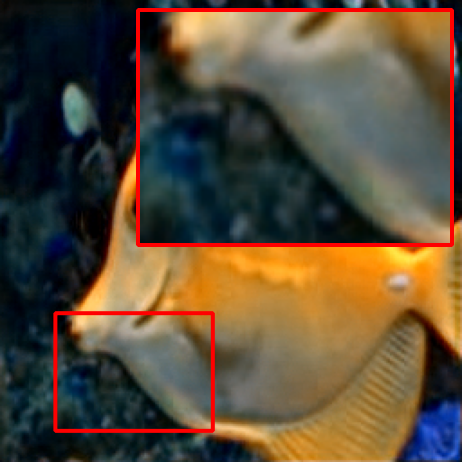}  \\
\rotatebox{90}{\textbf{KD}} &
\includegraphics[width=\imgs]{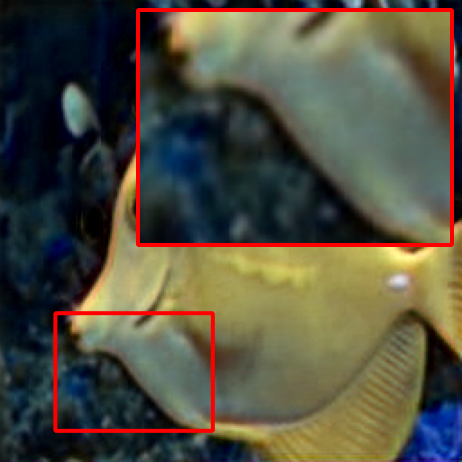} &
\includegraphics[width=\imgs]{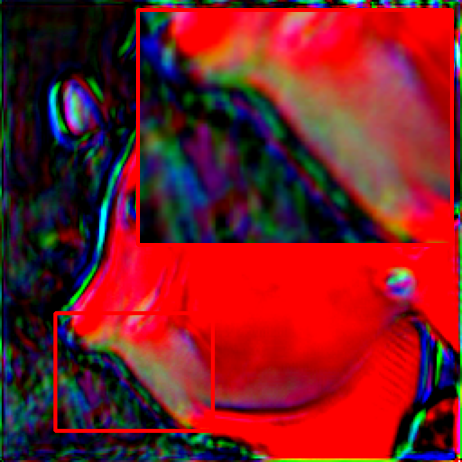} &
\includegraphics[width=\imgs]{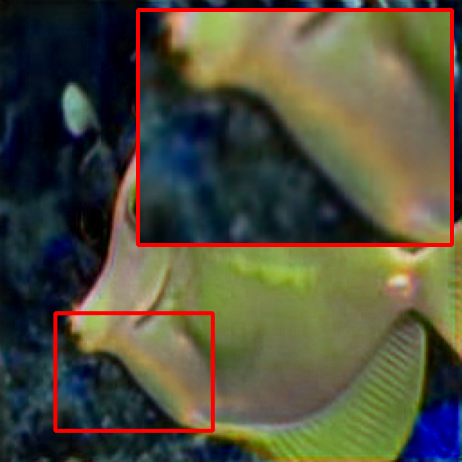} &
\includegraphics[width=\imgs]{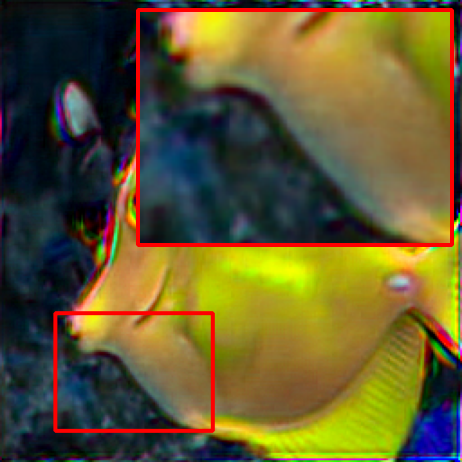} &
\includegraphics[width=\imgs]{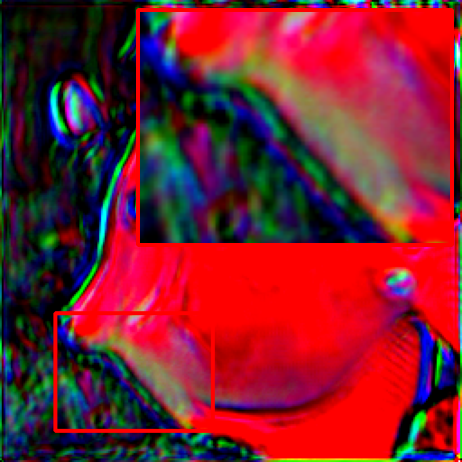} &
\includegraphics[width=\imgs]{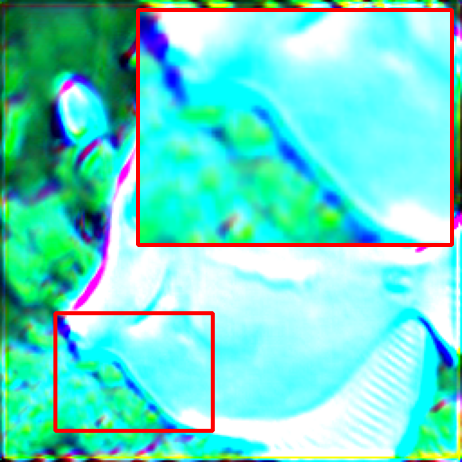} \\
& normal & noise & dropC & adv & ResNet-ASVP & Swin-ASVP \\
\end{tabular}
\end{center}\vspace{-2mm}
\caption{Underwater enhancement results. Our method degrades student predictions while retaining clear, natural outputs in the teacher.}

\vspace{-2mm}
\label{fig:UW}
\end{figure*}
%%%%%%%%%%%%%%%%%%% fig-3 %%%%%%%%%%%%%%%%%%%%%%%%%%%%%%%%%%%%%
%%%%%%%%%%%%%%%%%%% fig-4 %%%%%%%%%%%%%%%%%%%%%%%%%%%%%%%%%%%%%
\setlength{\tabcolsep}{1pt}
\def \imgl {0.14\linewidth}
\def \imgs {0.145\linewidth}
\begin{figure*}[!htb]
\small
\begin{center}
\begin{tabular}{ccccccc}
\rotatebox{90}{\textbf{Teacher}} &
\includegraphics[width=\imgs]{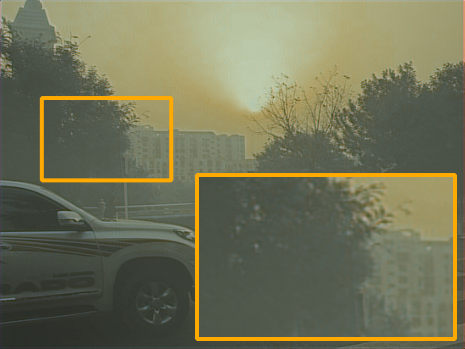} &
\includegraphics[width=\imgs]{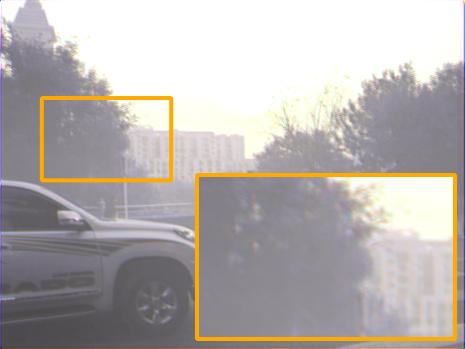} &
\includegraphics[width=\imgs]{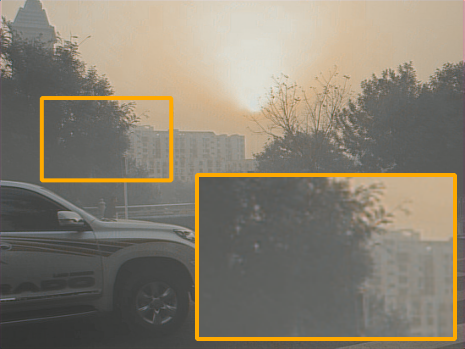} &
\includegraphics[width=\imgs]{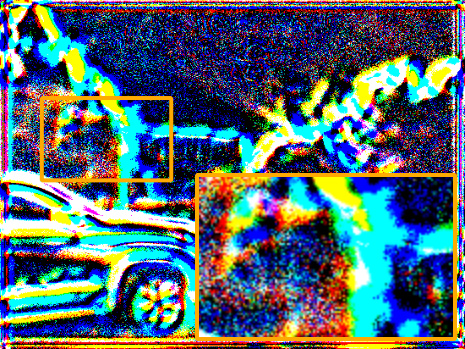} &
\includegraphics[width=\imgs]{fig/Dehazy/1/gt-svd_processed.png} & 
\includegraphics[width=\imgs]{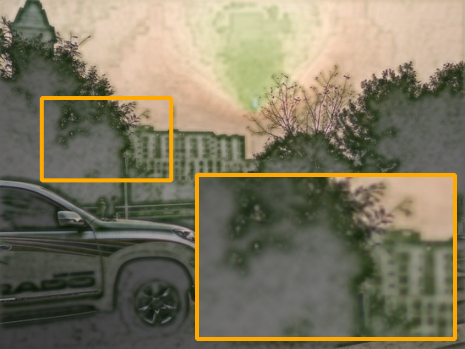}  \\
\rotatebox{90}{\textbf{KD}} &
\includegraphics[width=\imgs]{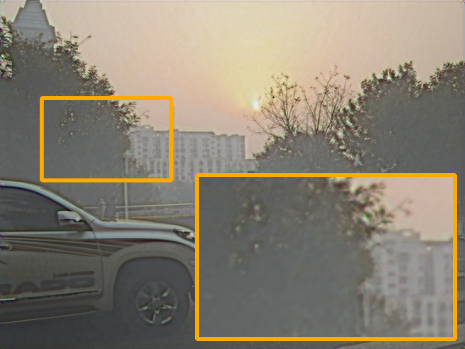} &
\includegraphics[width=\imgs]{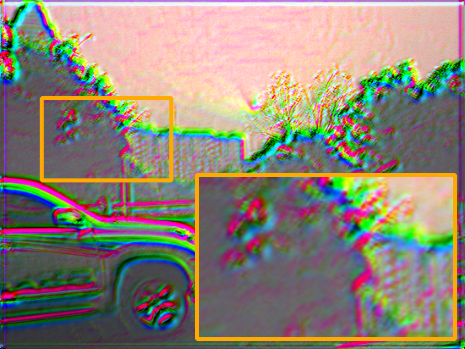} &
\includegraphics[width=\imgs]{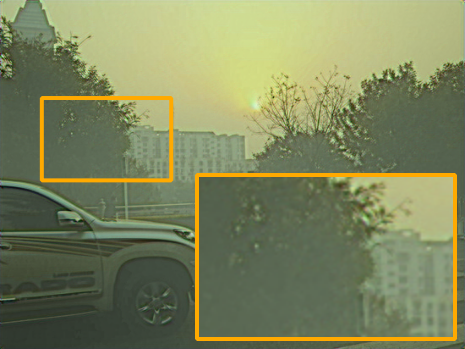} &
\includegraphics[width=\imgs]{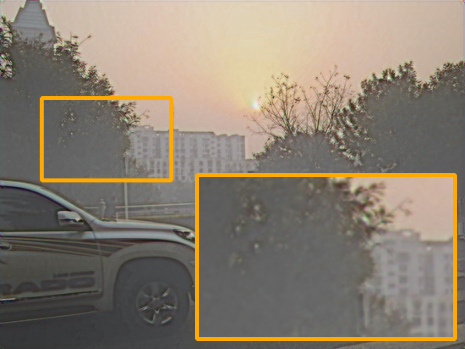} &
\includegraphics[width=\imgs]{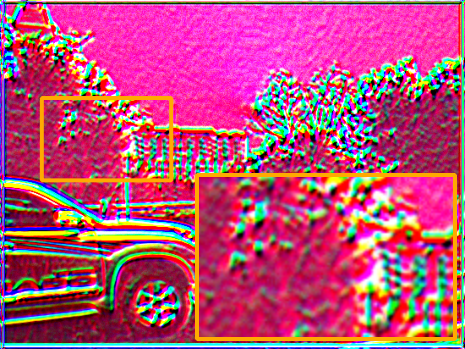} &
\includegraphics[width=\imgs]{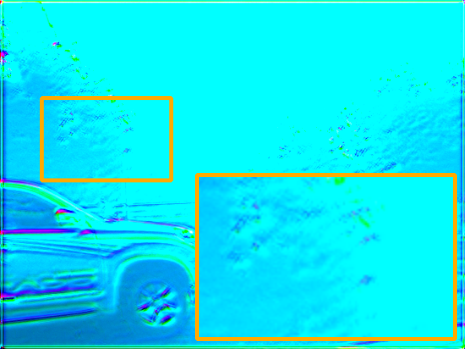} \\

\rotatebox{90}{\textbf{Teacher}} &
\includegraphics[width=\imgs]{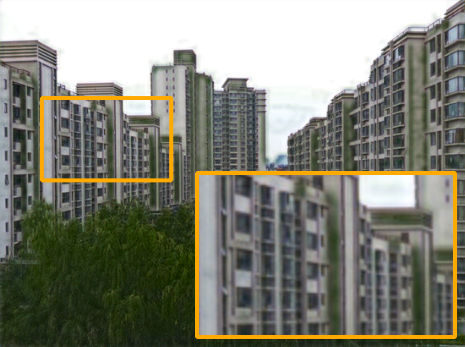} &
\includegraphics[width=\imgs]{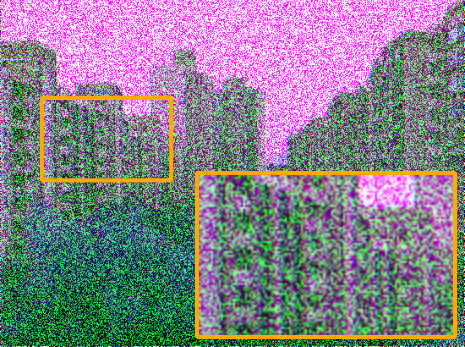} &
\includegraphics[width=\imgs]{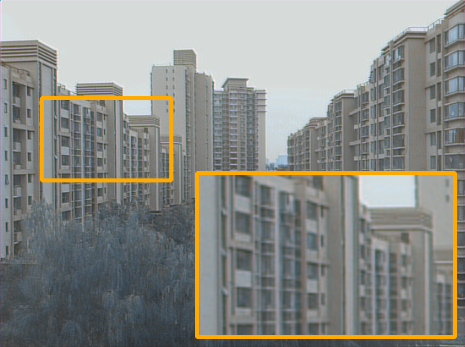} &
\includegraphics[width=\imgs]{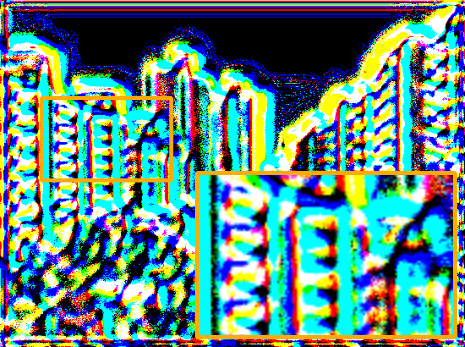} &
\includegraphics[width=\imgs]{fig/Dehazy/2/gt-svd_processed.png} & 
\includegraphics[width=\imgs]{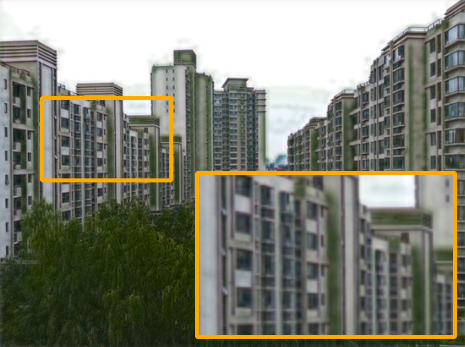}  \\
\rotatebox{90}{\textbf{KD}} &
\includegraphics[width=\imgs]{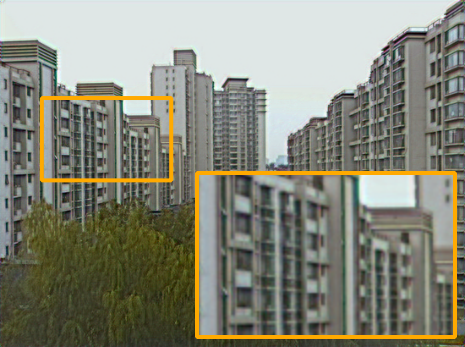} &
\includegraphics[width=\imgs]{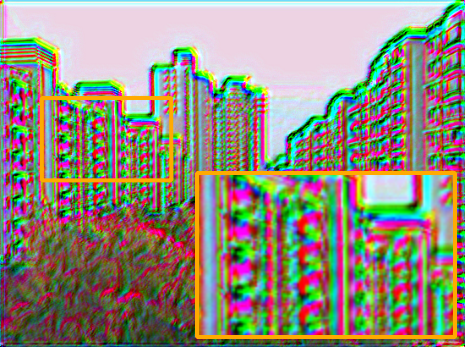} &
\includegraphics[width=\imgs]{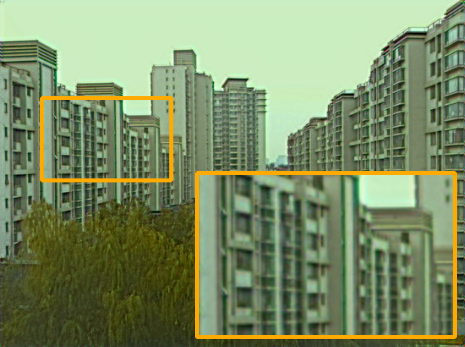} &
\includegraphics[width=\imgs]{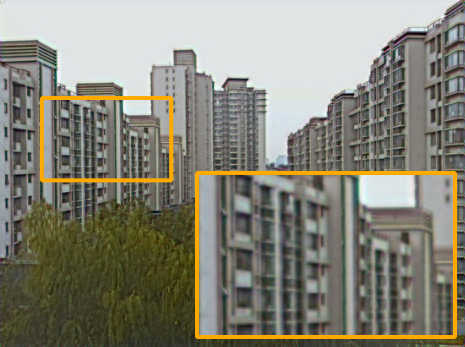} &
\includegraphics[width=\imgs]{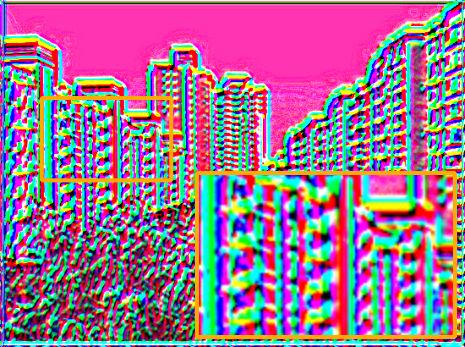} &
\includegraphics[width=\imgs]{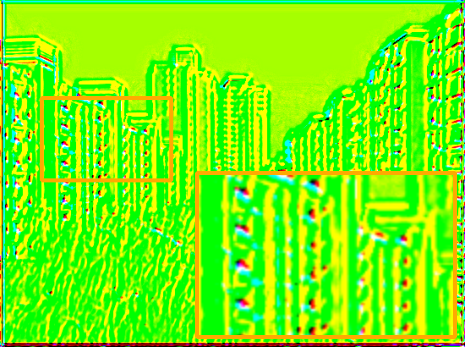} \\
& normal & noise & dropC & adv & ResNet-ASVP & Swin-ASVP \\
\end{tabular}
\end{center}\vspace{-2mm}
\caption{Dehazing results. Our method prevents students from recovering structural details, while teacher clarity is preserved.}

\vspace{-2mm}
\label{fig:Dehazy}
\end{figure*}
%%%%%%%%%%%%%%%%%%% fig-4 %%%%%%%%%%%%%%%%%%%%%%%%%%%%%%%%%%%%%

%%%%%%%%%%%%%%%%%%% fig-5 %%%%%%%%%%%%%%%%%%%%%%%%%%%%%%%%%%%%%
\setlength{\tabcolsep}{1pt}
\def \imgl {0.14\linewidth}
\def \imgs {0.15\linewidth}
\begin{figure*}[!htb]
\small
\begin{center}
\begin{tabular}{ccccccc}
\rotatebox{90}{\textbf{Teacher}} &
\includegraphics[width=\imgs]{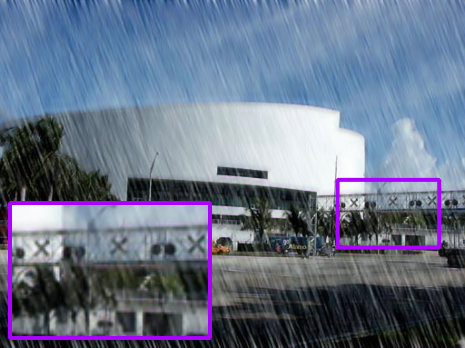} &
\includegraphics[width=\imgs]{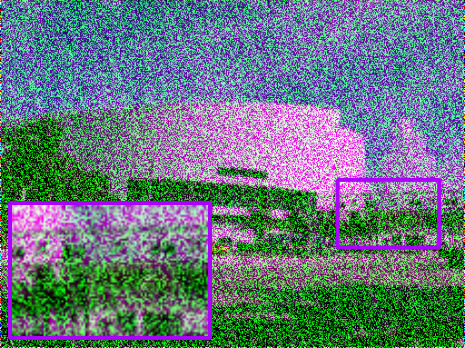} &
\includegraphics[width=\imgs]{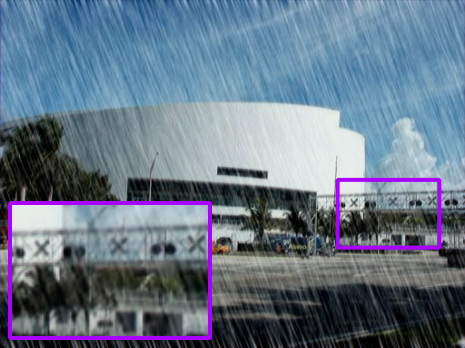} &
\includegraphics[width=\imgs]{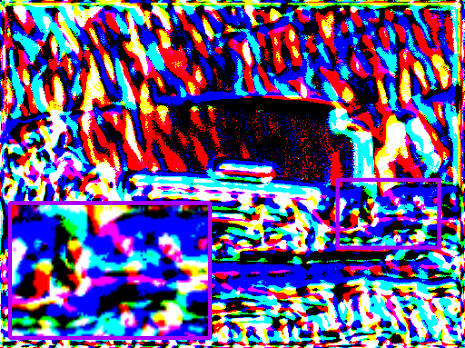} &
\includegraphics[width=\imgs]{fig/Derain/1/gt-svd_processed.png} & 
\includegraphics[width=\imgs]{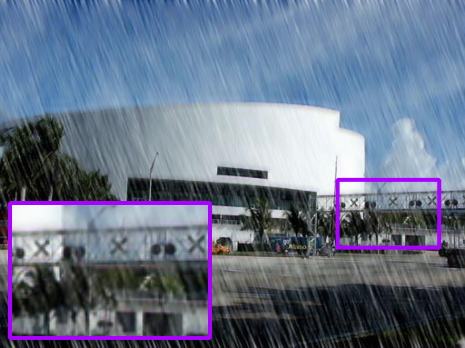}  \\
\rotatebox{90}{\textbf{KD}} &
\includegraphics[width=\imgs]{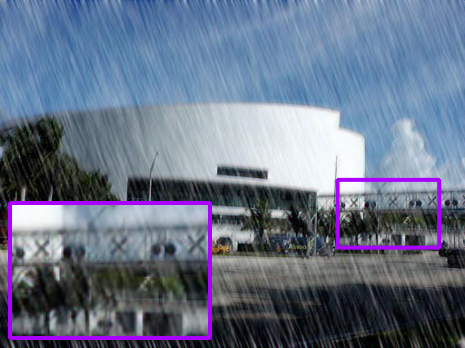} &
\includegraphics[width=\imgs]{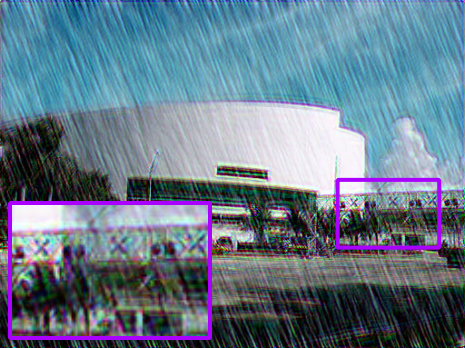} &
\includegraphics[width=\imgs]{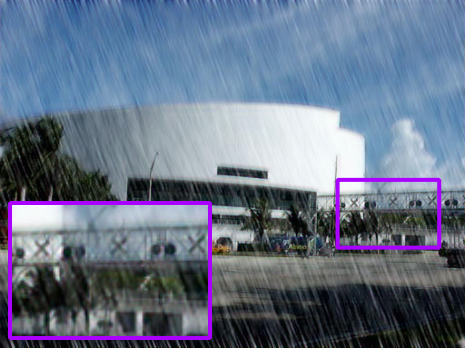} &
\includegraphics[width=\imgs]{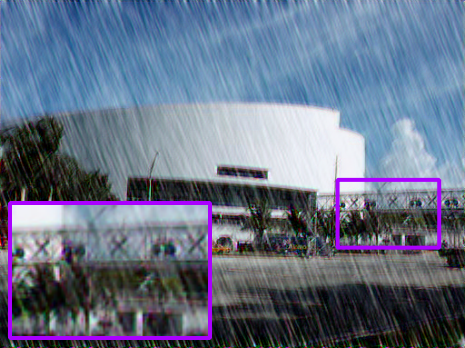} &
\includegraphics[width=\imgs]{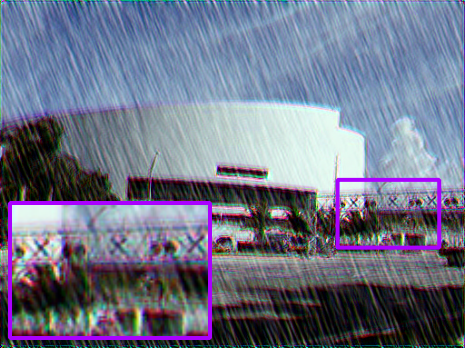} &
\includegraphics[width=\imgs]{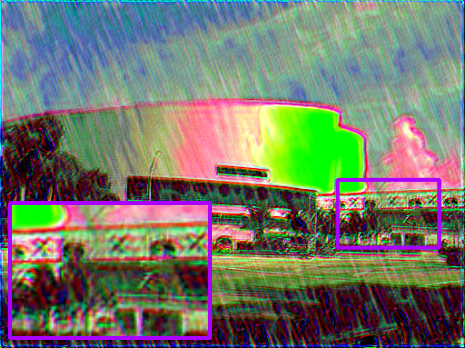} \\

\rotatebox{90}{\textbf{Teacher}} &
\includegraphics[width=\imgs]{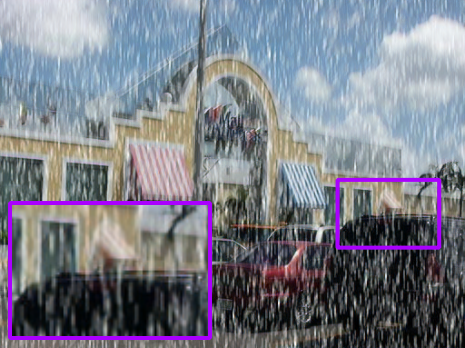} &
\includegraphics[width=\imgs]{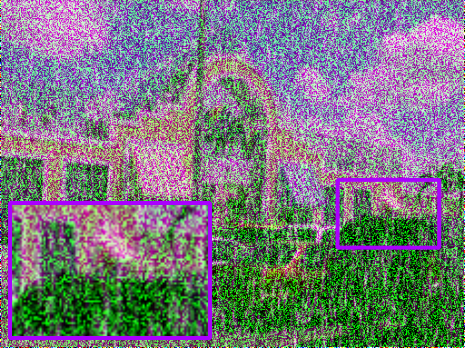} &
\includegraphics[width=\imgs]{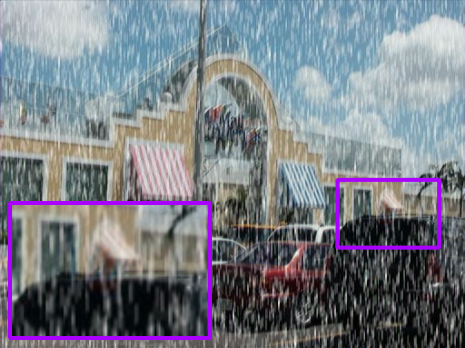} &
\includegraphics[width=\imgs]{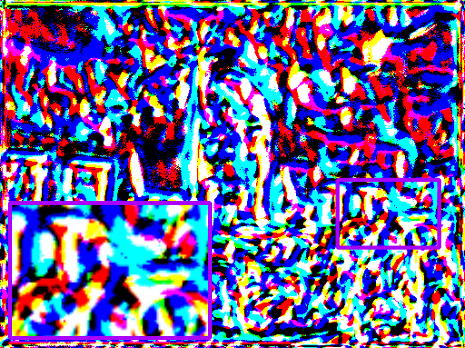} &
\includegraphics[width=\imgs]{fig/Derain/255/gt-svd_processed.png} & 
\includegraphics[width=\imgs]{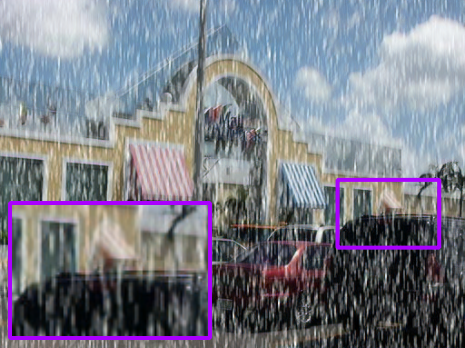}  \\
\rotatebox{90}{\textbf{KD}} &
\includegraphics[width=\imgs]{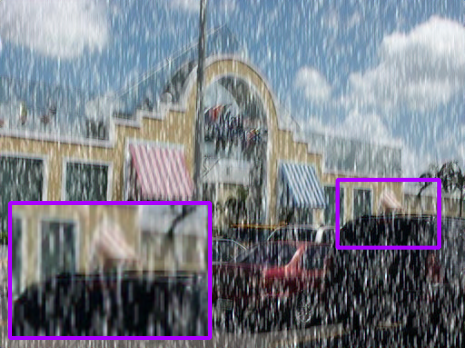} &
\includegraphics[width=\imgs]{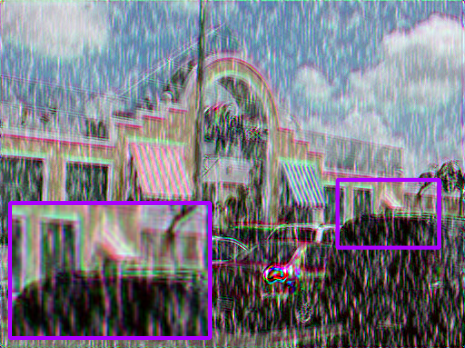} &
\includegraphics[width=\imgs]{fig/Derain/255/drop_processed.png} &
\includegraphics[width=\imgs]{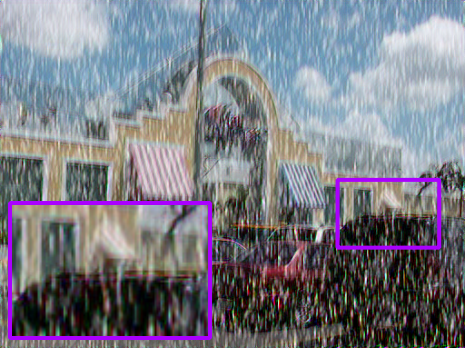} &
\includegraphics[width=\imgs]{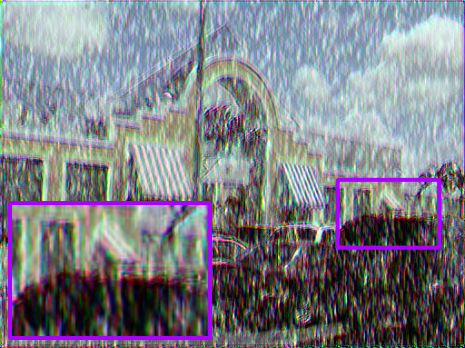} &
\includegraphics[width=\imgs]{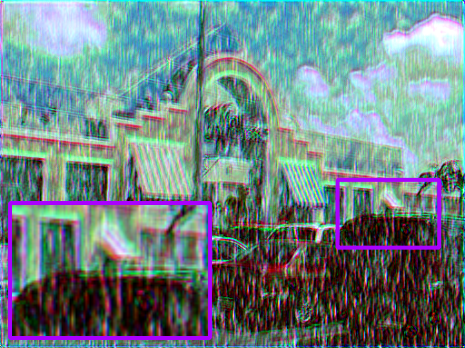} \\
& normal & noise & dropC & adv & ResNet-ASVP & Swin-ASVP \\
\end{tabular}
\end{center}\vspace{-2mm}
\caption{Deraining results. Our method disrupts student restoration quality, with the teacher output remaining sharp and rain-free.}

\vspace{-2mm}
\label{fig:Derain}
\end{figure*}
%%%%%%%%%%%%%%%%%%% fig-5 %%%%%%%%%%%%%%%%%%%%%%%%%%%%%%%%%%%%%
\subsection{Performance Evaluation}
\noindent\textbf{Image Super Resolution Results.} Table.~\ref{tab:SR} shows super resolution results on DIV2K\cite{agustsson2017ntire_DIV2k}. Without defense, both ResNet-18 and SwinIR teachers attain PSNR/SSIM around 32 dB / 0.87, and student ResNet-9/ResNet-18 achieve nearly the same metrics, indicating effective knowledge transfer. With ASVP, the teachers maintain about the same PSNR/SSIM, but the student PSNR drops by up to 4 dB and SSIM drops dramatically (by 60-75\%). This gap reflects that the student output, while roughly similar in mean squared error (affecting PSNR), has lost structural detail (affecting SSIM). Qualitatively, the ASVP induced student reconstructions exhibit high frequency artifacts and blurring of fine details in Fig.~\ref{fig:SR}: edges and textures are less clear, as the student is effectively “chasing” noisy spectral components. In contrast, the teacher outputs remain sharp and faithful to the ground truth, since the model counteracts the injected perturbations internally. Thus, in super resolution, ASVP preserves fidelity in the teacher (high PSNR/SSIM) but disrupts the student’s ability to capture fine grained textures, leading to lower scores.

% We evaluate our defense method on the DIV2K dataset\cite{agustsson2017ntire_DIV2k}. As summarized in Table.~\ref{tab:SR}, the proposed approach significantly suppresses student model performance across all architectures, while maintaining the original quality of the teacher outputs. Fig.~\ref{fig:SR} presents the visual results under various distillation defenses. It can be clearly observed that, compared to noise, dropout, and adversarial perturbations, our method introduces structured degradation in the student results, while preserving high-frequency textures in the teacher reconstructions.

\noindent\textbf{Low Light Enhancement Results.} Table.~\ref{tab:LLM}  summarizes low light enhancement results on LOLv1\cite{wei2018deep_LOL}. The ResNet-18 teacher brightens the low light images effectively, and the student network under baseline KD learns a similar mapping, resulting in only modest performance loss. After ASVP is enabled, the teacher’s output brightness and detail are essentially preserved, but the student’s PSNR/SSIM degrade notably. The student’s enhanced images under ASVP show over amplified noise in dark regions and inconsistent illumination in Fig.~\ref{fig:LLM}: details in shadows are lost and color fidelity is reduced. These artifacts arise because the ASVP perturbations add spectral components that are “rolled back” by the teacher’s layers, but the student has difficulty reproducing consistent luminance adjustment. Quantitatively, this manifests as a substantial SSIM drop and a several dB PSNR decline. The contrast and edges  are no longer well captured by the student, indicating structural information loss due to the injected perturbations.

% Table.~\ref{tab:LLM} shows the performance on the LOL-v1 dataset\cite{wei2018deep_LOL} under different defense settings. Our method consistently achieves strong undistillation effectiveness without compromising teacher image quality. As depicted in Fig.~\ref{fig:LLM}, traditional defenses leave the student outputs relatively clean or slightly distorted. In contrast, the proposed method causes severe degradation in student reconstructions, while the teacher retains detail and natural lighting consistency.

\noindent\textbf{Underwater Image Enhancement Results.}Table.~\ref{tab:UW} presents underwater image enhancement on LSUI\cite{peng2023u——LSUI}. Underwater images feature color casts and haze; the SwinIR teacher removes color shifts and restores clarity, and the student without defense achieves similar performance. With ASVP, again the teacher’s outputs remain clear, but student reconstructions are marred by residual color noise and loss of contrast. Visually, ASVP introduces high frequency speckles and incorrect color patches in the student output, undermining structural similarity. This noise causes the student’s SSIM to drop much more than PSNR: PSNR is somewhat robust to small pixel deviations, whereas SSIM is sensitive to the overall color consistency and contrast structure being broken. As shown in Fig.~\ref{fig:UW}, the student’s SSIM plummets, signaling that meaningful content is distorted. These qualitative distortions reflect the ASVP mechanism at work: by amplifying dominant singular values, the teacher’s layers conceal this distortion, but the student, learning from these misleading features, produces inaccurate texture and color predictions.

%%%%%%%%%%%%%%%%%%% vali %%%%%%%%%%%%%%%%%%%%%%%%%%%%%%%%%%%%%
\setlength{\tabcolsep}{1pt}
\def \imgb {0.17\linewidth}
\def \imgs {0.16\linewidth}
\begin{figure*}[!htb]
\small
\begin{center}
\begin{tabular}{ccccccc}

\rotatebox{90}{\textbf{features}} &
\includegraphics[width=\imgs]{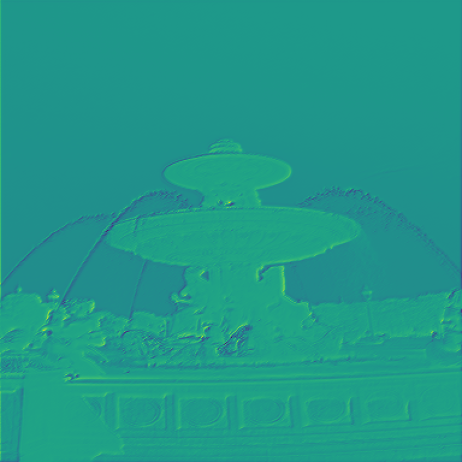} &
\includegraphics[width=\imgs]{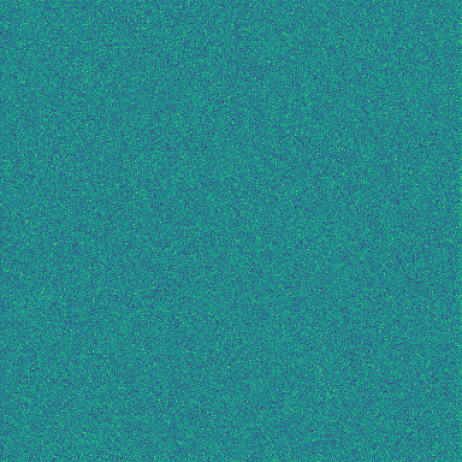} &
\includegraphics[width=\imgs]{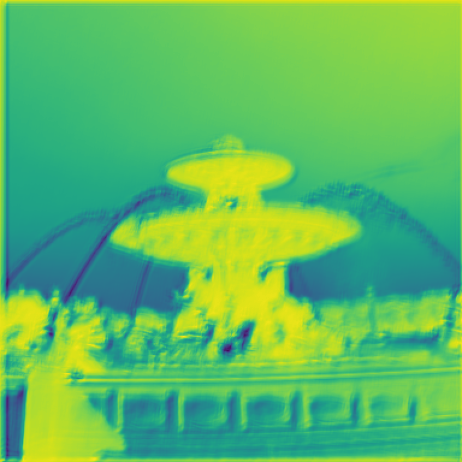} &
\includegraphics[width=\imgs]{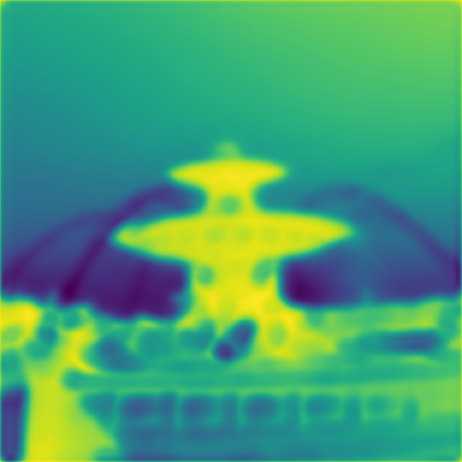} &
\includegraphics[width=\imgs]{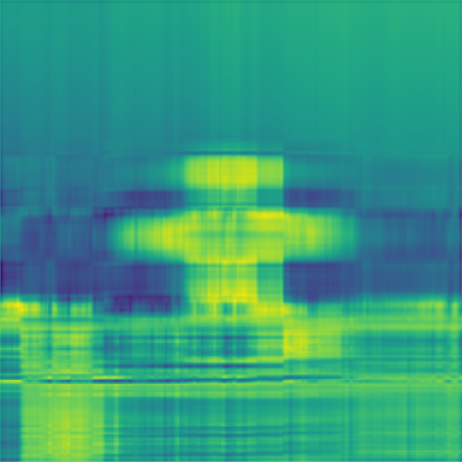} &   \\
\rotatebox{90}{\textbf{energy}} &
\includegraphics[width=\imgb]{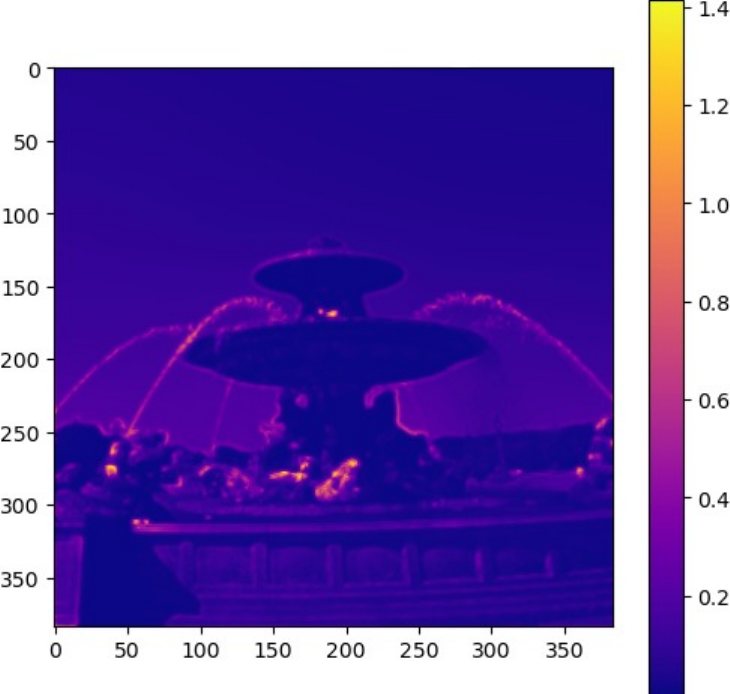} &
\includegraphics[width=\imgb]{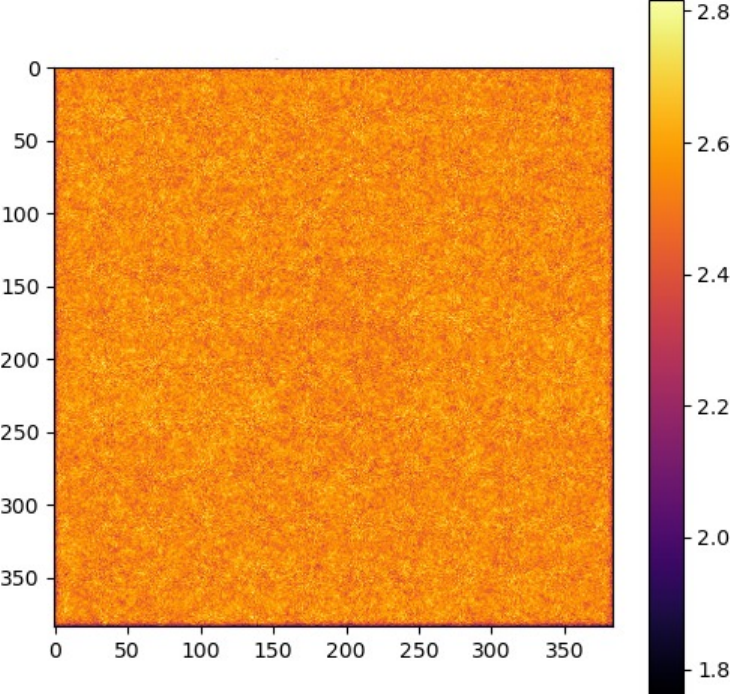} &
\includegraphics[width=\imgs]{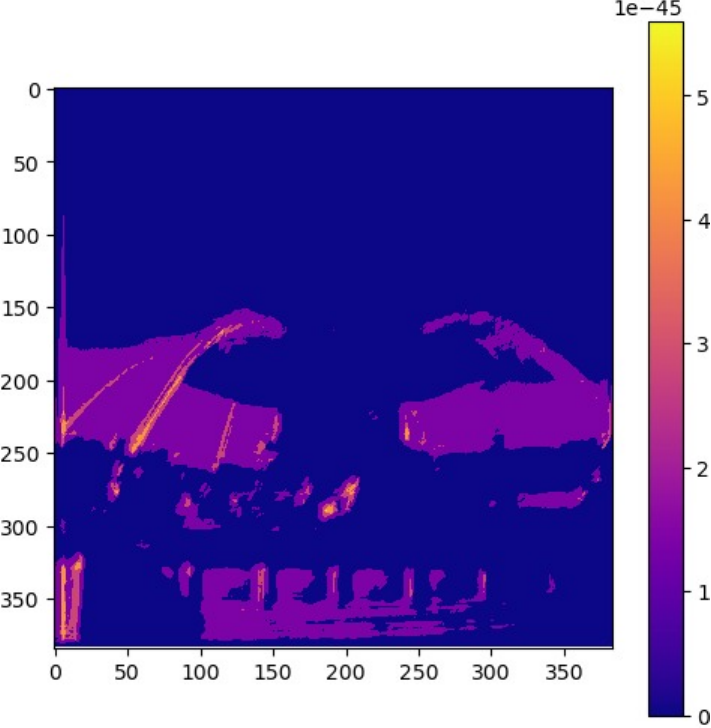} &
\includegraphics[width=\imgb]{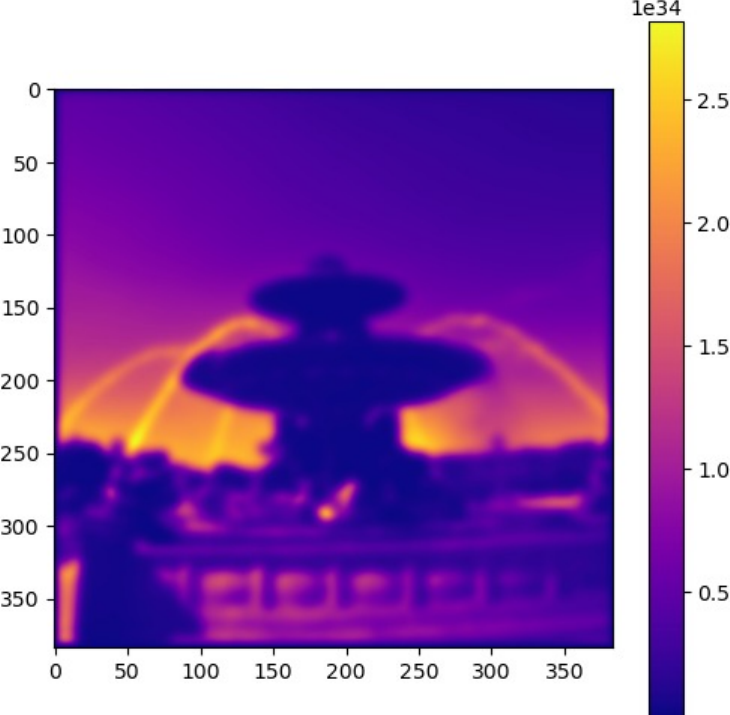} &
\includegraphics[width=\imgb]{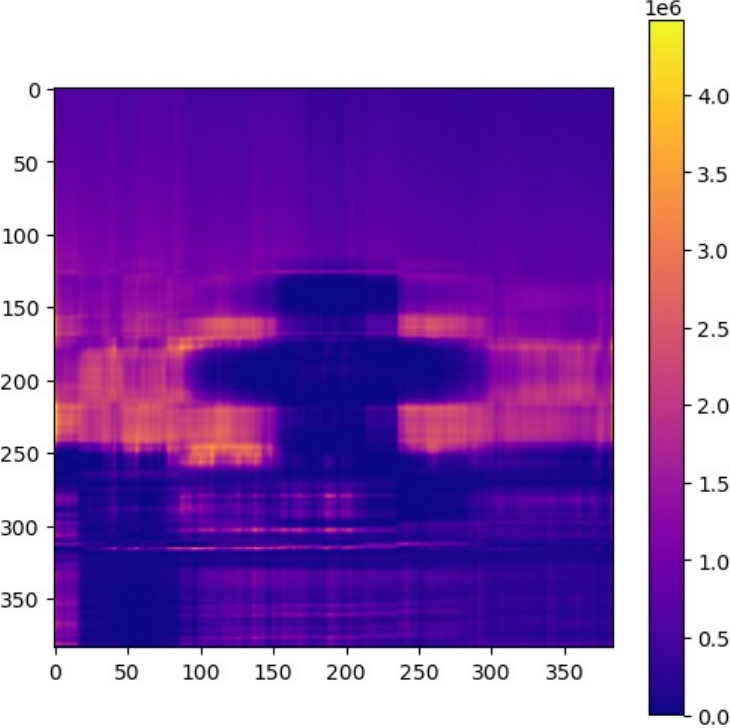} & \\

\rotatebox{90}{\textbf{frequency}} &
\includegraphics[width=\imgb]{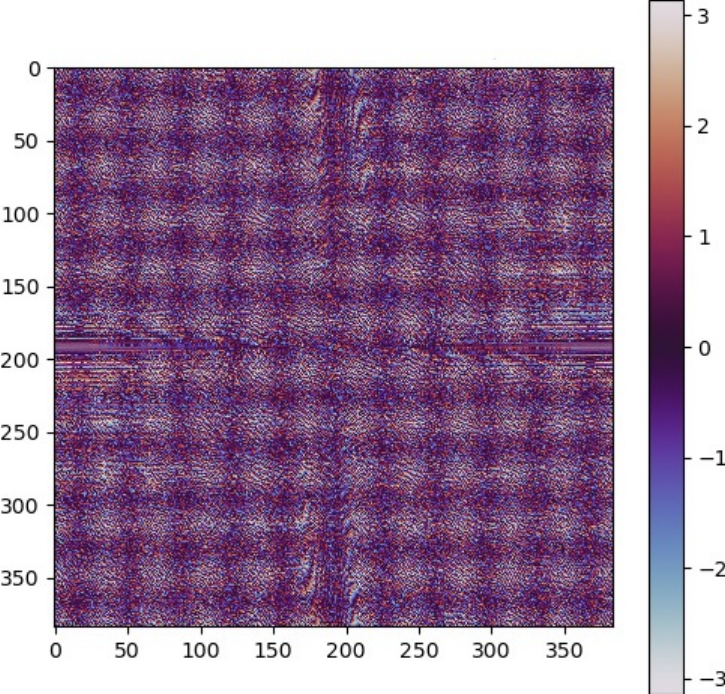} &
\includegraphics[width=\imgb]{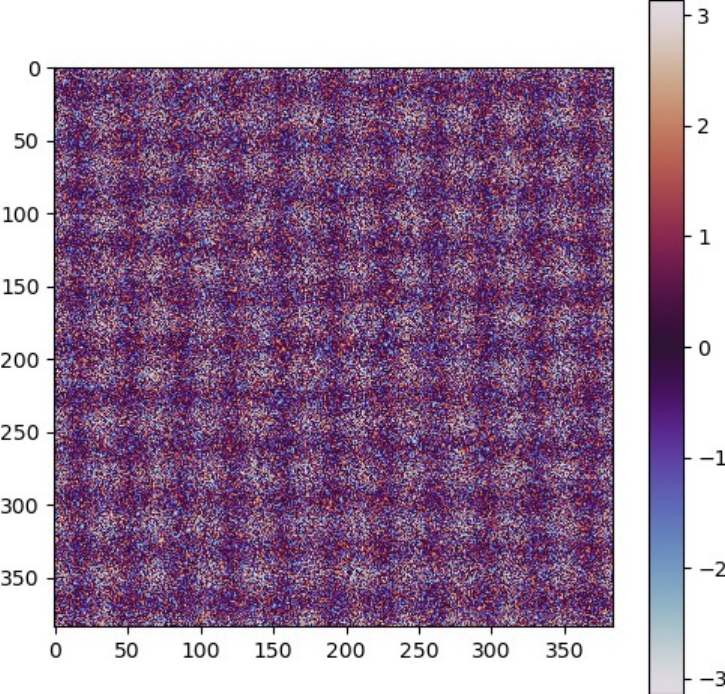} &
\includegraphics[width=\imgb]{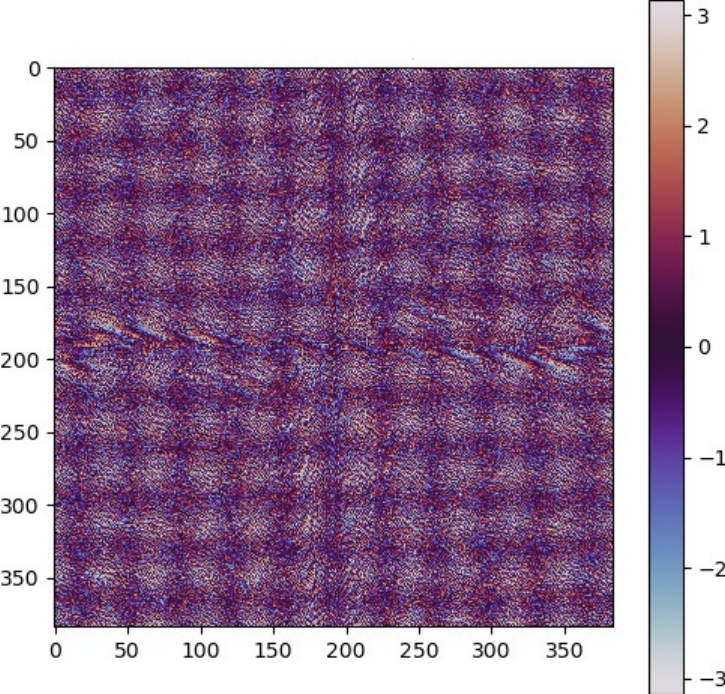} &
\includegraphics[width=\imgb]{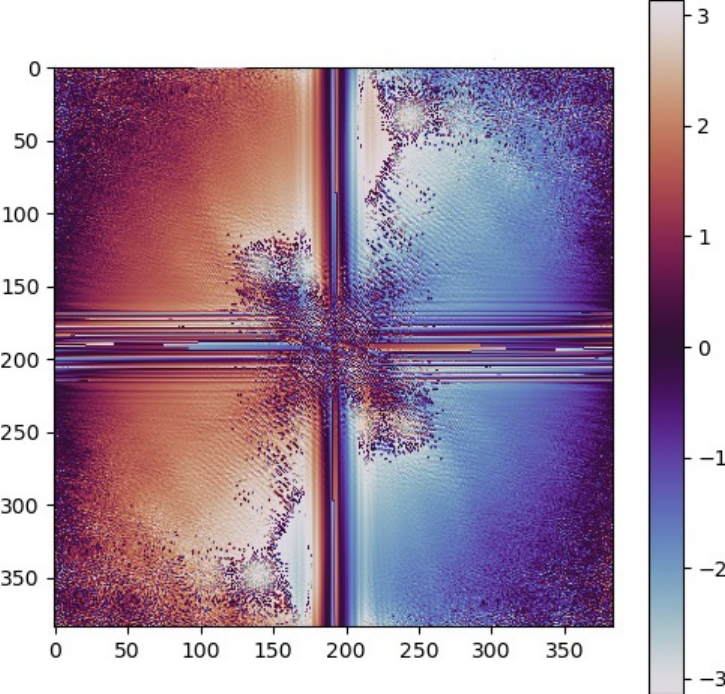} &
\includegraphics[width=\imgb]{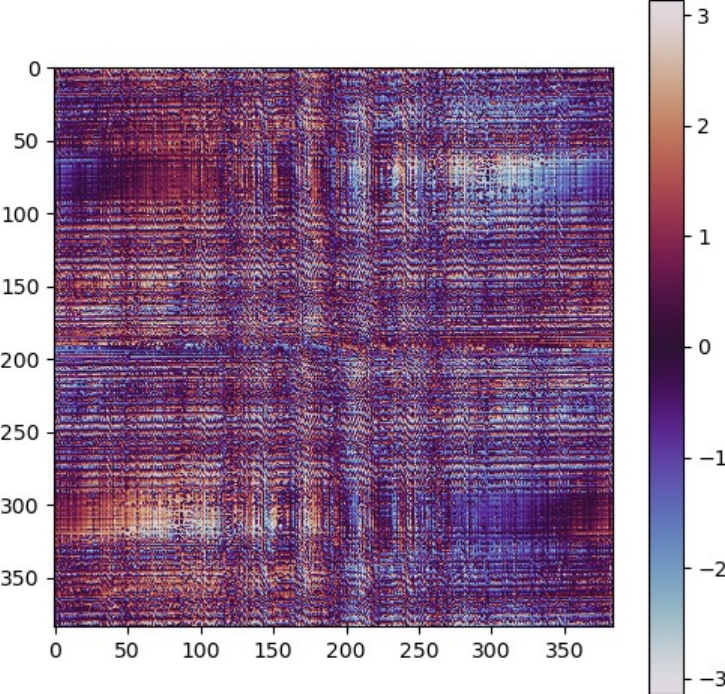} &   \\

\rotatebox{90}{\textbf{numerical}} &
\includegraphics[width=\imgb]{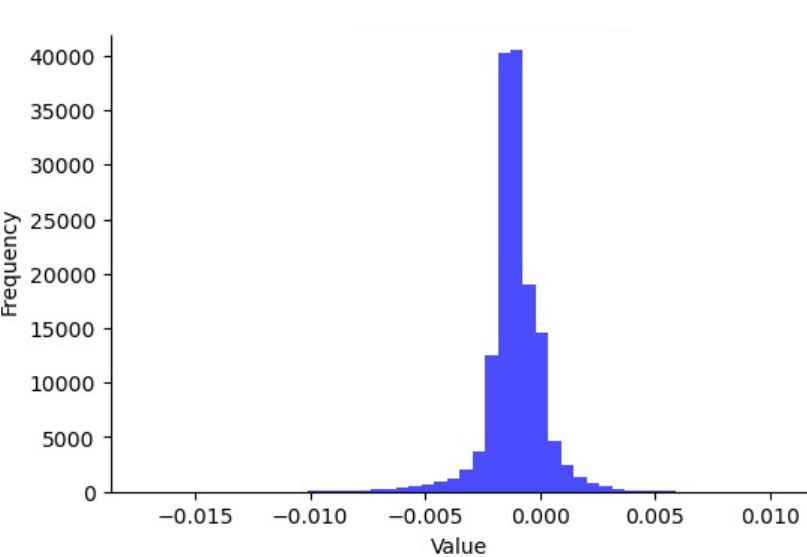} &
\includegraphics[width=\imgb]{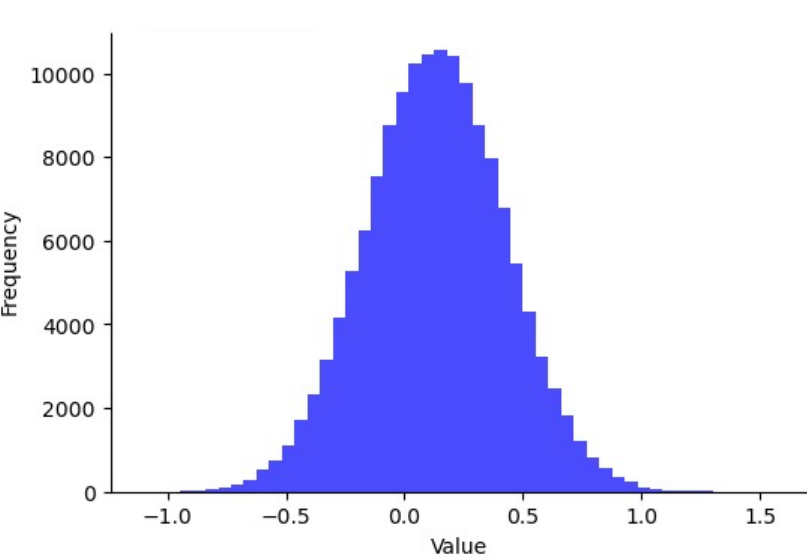} &
\includegraphics[width=\imgb]{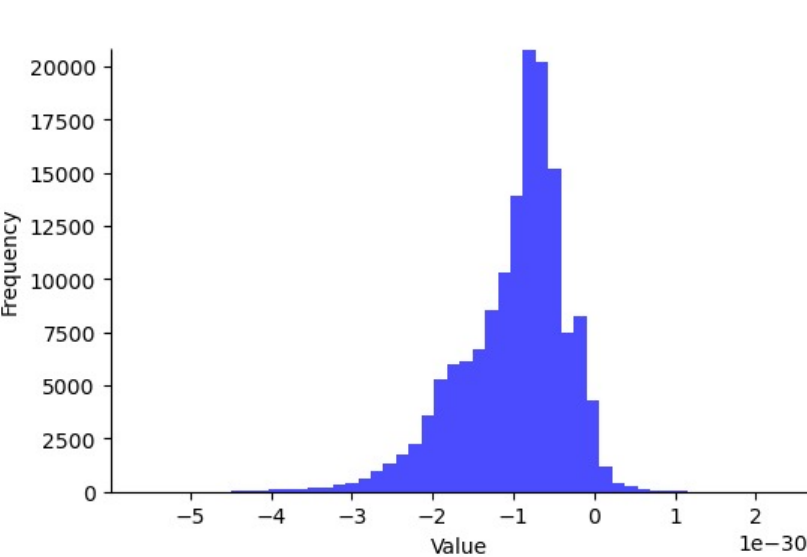} &
\includegraphics[width=\imgb]{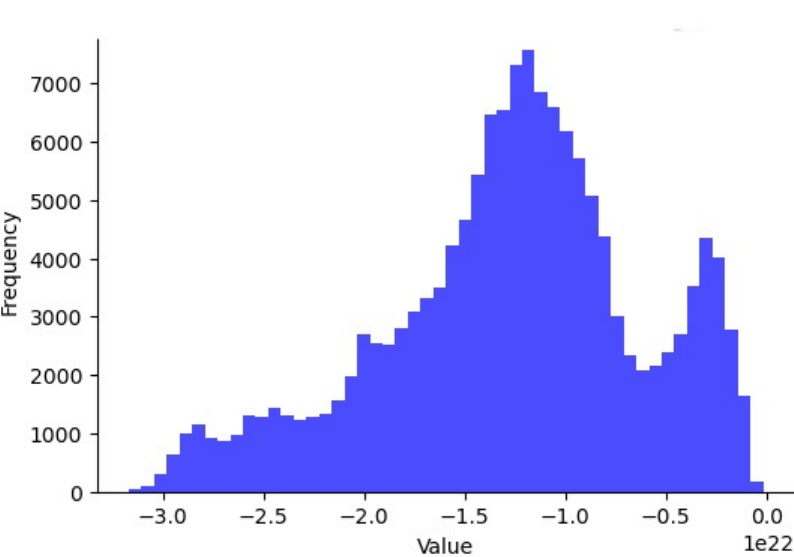} &
\includegraphics[width=\imgb]{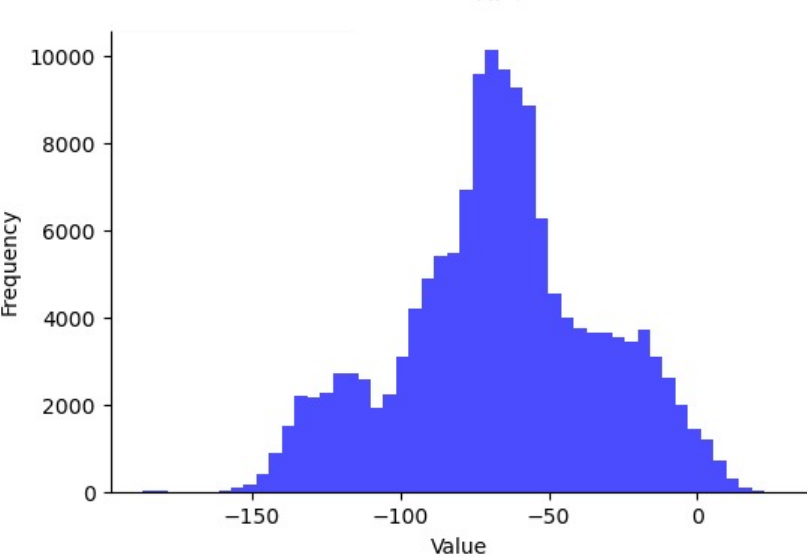} &   \\

\rotatebox{90}{\textbf{3D}} &
\includegraphics[width=\imgs]{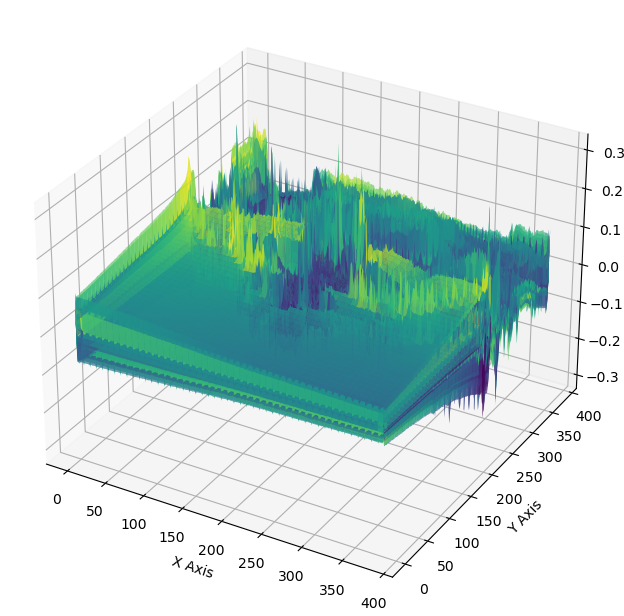} &
\includegraphics[width=\imgs]{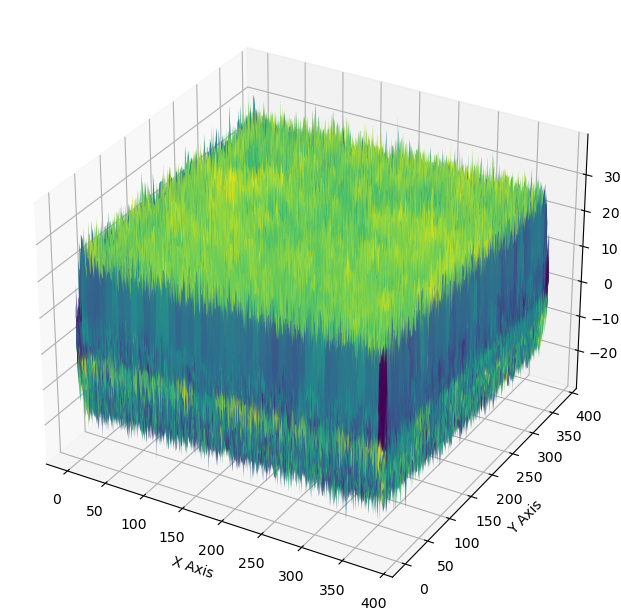} &
\includegraphics[width=\imgs]{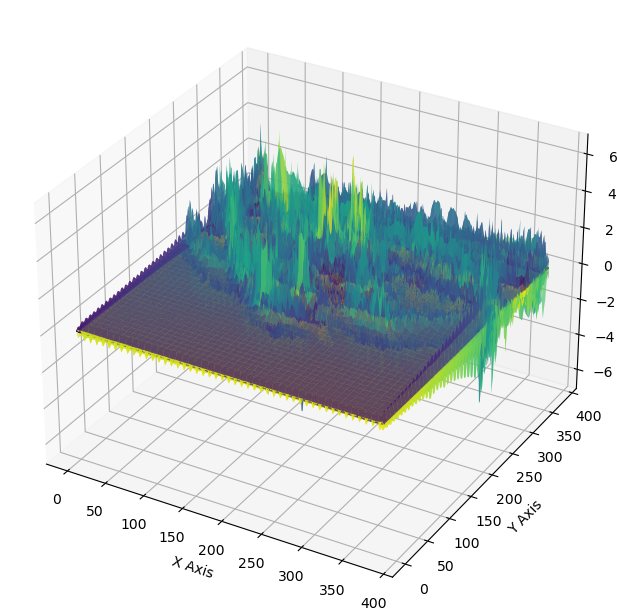} &
\includegraphics[width=\imgs]{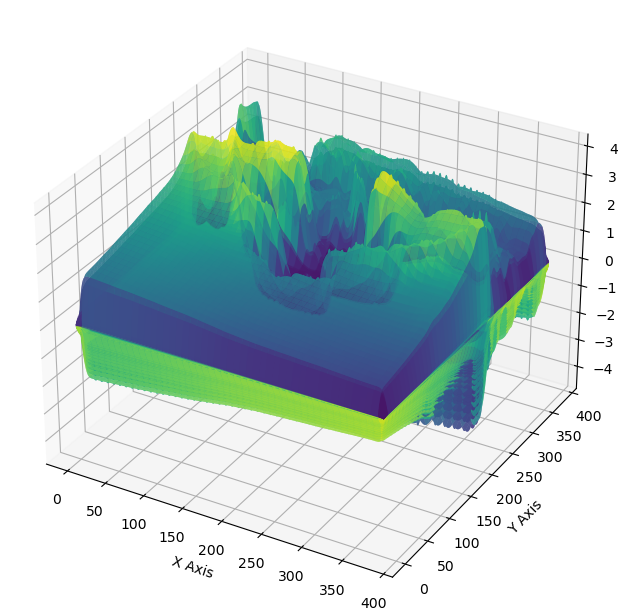} &
\includegraphics[width=\imgs]{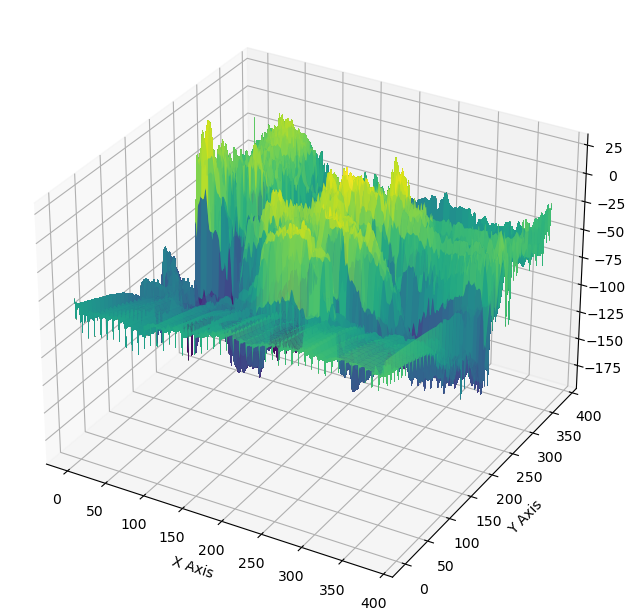} &   \\

& normal & noise & dropC & adv & ResNet-ASVP  \\
\end{tabular}
\end{center}\vspace{-2mm}
\caption{Multi-dimensional analysis of feature distortions. Our method introduces spatial, spectral, and numerical disruptions, effectively impeding student learning.}

\vspace{-2mm}
\label{fig:vali}
\end{figure*}
%%%%%%%%%%%%%%%%%%% vali %%%%%%%%%%%%%%%%%%%%%%%%%%%%%%%%%%%%%
%%%%%%%%%%%%%%%%%%% k and h %%%%%%%%%%%%%%%%%%%%%%%%%%%%%%%%%%%%%
\begin{figure}[!h]
    \centering
    
    \includegraphics[width=0.8\linewidth]{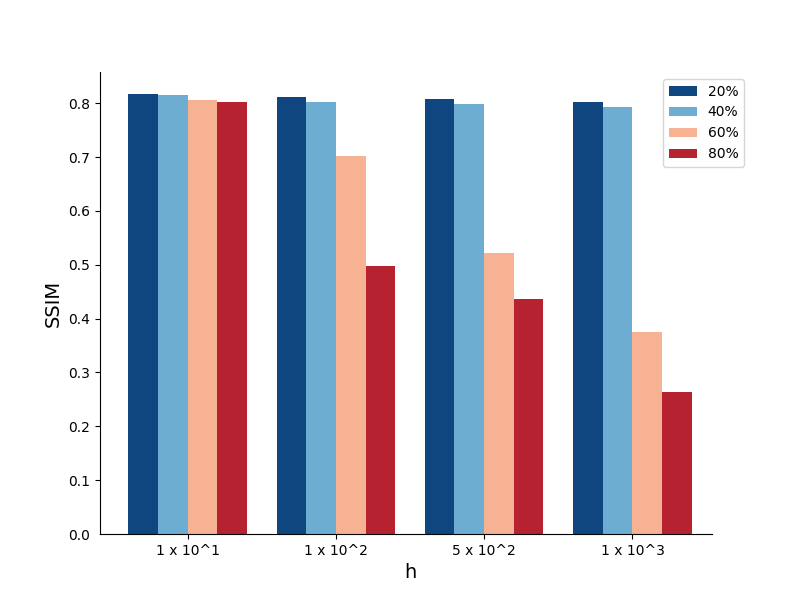}
    \caption{SSIM scores of the student model under different ASVP parameter settings on the dehazing dataset.$k$ indicates the percentage of top singular values amplified, and $h$ is the amplification factor. For reference, the SSIM of the student model under standard distillation (without ASVP) is 0.820. Lower SSIM indicates stronger defense effectiveness.
}

    \label{fig:kh}
\end{figure}
%%%%%%%%%%%%%%%%%%% k and h %%%%%%%%%%%%%%%%%%%%%%%%%%%%%%%%%%%%%

% Quantitative results are provided in Table.~\ref{tab:UW}. Our method substantially reduces student SSIM and PSNR across all model variants. Fig.~\ref{fig:UW} illustrates visual comparisons on challenging underwater scenes. The student model trained under our defense fails to recover structure and contrast, demonstrating the defense's robustness, whereas the teacher produces clear, vibrant underwater imagery.

\noindent\textbf{Image Dehazing Results.} Table.~\ref{tab:Dehaze} shows dehazing results on the SOTS \cite{liu2019learning_SOTS}. The ResNet-18 teacher effectively removes fog and recovers scene contrast, achieving high scores. Under normal KD, students perform well in removing haze. With ASVP, the teacher’s clean images are maintained, but student outputs retain noticeable haze or introduce halos around edges. In other words, ASVP causes the student to under enhance the image: PSNR falls by multiple dB and SSIM drops significantly. The distortion here is structural in Fig.~\ref{fig:Dehazy}: true scene details  remain blurred in the student’s image. The amplified singular components injected by ASVP seem to disrupt the teacher’s feature alignment, so the student fails to learn the correct global contrast mapping. Again, the teacher compensates for these perturbations internally, whereas the student exhibits decreased detail recovery and higher global error, as reflected in the lowered metrics.

% As shown in Table.~\ref{tab:Dehaze}, our method outperforms existing defenses in both preserving teacher performance and weakening student predictions on the SOTS dataset\cite{liu2019learning_SOTS}. In Fig.~\ref{fig:Dehazy}, visual comparisons show that our approach maintains sharp dehazed results in the teacher model, while the student reconstructions exhibit significant blur and residual haze, unlike other methods that retain partial structure.

\noindent\textbf{Image Deraining Results.} Table.~\ref{tab:Derain} reports deraining results on Rain100\cite{yang2017deep_Rain}. Both teacher models remove rain streaks effectively under normal conditions. With ASVP, the teachers’ results are almost unchanged, but student outputs often retain fine streaks or introduce random artifacts resembling noise. This results in a clear quantitative gap: the student’s PSNR/SSIM under ASVP is much lower than baseline KD. Visually, in Fig.~\ref{fig:Derain}, the ASVP student outputs appear as if some rain patterns or speckle noise persist on the image, indicating a loss of clean structure. The singular value perturbations have injected spurious high frequency patterns, which the student incorrectly interprets as part of the scene texture. Because SSIM penalizes such structural discrepancies, the student’s SSIM score suffers greatly, even if the mean signal level  is somewhat similar. These results confirm that ASVP effectively prevents the student from aligning with the teacher’s internal representations, so the student cannot fully remove the rain despite training on the teacher’s outputs.

% Table.~\ref{tab:Derain} summarizes results on the Rain100 dataset\cite{yang2017deep_Rain}. Our defense mechanism again achieves the best trade-off between teacher stability and student degradation. Fig.~\ref{fig:Derain} visually compares deraining outputs across different strategies. Competing defenses fail to completely remove streaks in the student results. In contrast, our method successfully disrupts student performance while preserving detailed restoration in the teacher output.

%%%%%%%%%%%%%%%%%%%%%%%%%%%%%%%%%%%%%%%%%%%%%%%5.1%%%%%%%%%%%%%%%%%%%%%%%%%%%%%%%%%
\begin{table}[htbp]
\centering
\caption{Student Model Performance under KD from Different Perturbed Feature Stages (Teacher with Full-Layer ASVP Applied). ASVP is applied to all teacher layers. The student selects different feature stages (early, mid, late, or all) for distillation. Results shown for super-resolution on DIV2K.}
\label{tab:distill_stage}
\begin{tabular}{p{5.5cm} p{2.5cm} p{2.5cm} p{2.5cm} }
\toprule
\textbf{Distillation Feature Stage} & \textbf{PSNR (↑)} & \textbf{SSIM (↑)} & \textbf{LPIPS (↓)}  \\
\midrule
None (clean KD)            & \textbf{31.98 dB} & \textbf{0.838}    & \textbf{0.157}      \\
Early-stage features       & 29.12 dB          & 0.744             & 0.282               \\
Mid-stage features         & 28.45 dB          & 0.736             & 0.243               \\
Late-stage features        & 27.91 dB          & 0.709             & 0.288               \\
\bottomrule
\end{tabular}
\end{table}

%%%%%%%%%%%%%%%%%%%%%%%%%%%%%%%%%%%%%%%%%%%%%%%5.1%%%%%%%%%%%%%%%%%%%%%%%%%%%%%%%%%

\subsection{Feature Level Impact Analysis}
To better understand how different defense strategies influence the knowledge transfer process, we conduct a detailed comparison of intermediate feature maps from various teacher models including baseline (normal), noise injected, channel dropout, adversarial, and our proposed defense—along five analytical dimensions: visual appearance, energy distribution, frequency content, numerical activation spread, and 3D spatial structure in Fig~\ref{fig:vali}.

\subsubsection{Feature Map Visualization}
Feature maps from the normal model and the channel dropout model are sharp and stable, preserving structural integrity effectively. Although adversarial perturbations introduce local distortions, the overall structure remains intact. In comparison, noise perturbations inject randomness into fine grained details, yet a degree of global structural information is preserved, allowing the student model to still learn useful patterns. However, under our defense, the student model is no longer able to replicate key structures. The proposed ASVP module introduces structured periodic fluctuations by amplifying singular values, making feature maps unstable and inconsistent. As a result, it significantly hinders the student’s ability to extract meaningful representations from the teacher.

%%%%%%%%%%%%%%%%%%%%%%%%%%%%%%%%%%%%%%%%%%%%%%%5.2%%%%%%%%%%%%%%%%%%%%%%%%%%%%%%%%%
\begin{table}[htbp]
\centering
\caption{Minimum effective hyperparameters for ASVP defense across five restoration tasks. Defense is considered effective when the student model suffers a PSNR drop of at least 1.5 dB, or an SSIM degradation of 0.1 or more.}
\label{tab:min_hk}
\begin{tabular}{l>{\centering\arraybackslash}p{4.2cm}>{\centering\arraybackslash}p{4cm}}
\toprule
\textbf{Task} & \textbf{Min Amplification Factor $h$} & \textbf{Min Top-$k$ Ratio (\%)} \\
\midrule
Super-resolution         & $1 \times 10^2$ & 40\% \\
Low-light Enhancement   & $1 \times 10^2$ &40\% \\
Dehazing                 & $1 \times 10^2$ &60\% \\
Underwater Enhancement  & $1 \times 10^2$ &60\% \\
Deraining               & $1 \times 10^2$ &60\% \\
\bottomrule
\end{tabular}
\end{table}

%%%%%%%%%%%%%%%%%%%%%%%%%%%%%%%%%%%%%%%%%%%%%%%5.2%%%%%%%%%%%%%%%%%%%%%%%%%%%%%%%%%

%%%%%%%%%%%%%%%%%%%%%%%%%%%%%%%%%%%%%%%%%%%%%%%5.3%%%%%%%%%%%%%%%%%%%%%%%%%%%%%%%%%
\begin{table}[htbp]
\centering
\caption{Runtime and memory comparison of different defense methods applied to teacher model (ResNet18). Values measured on NVIDIA RTX 3090 using 256×256 and 4K (3840×2160) inputs. Adversarial perturbation implemented via 3-step PGD.}
\label{tab:runtime_overhead}
\begin{tabular}{l>{\centering\arraybackslash}p{2.2cm}>{\centering\arraybackslash}p{2cm}>{\centering\arraybackslash}p{2.2cm}>{\centering\arraybackslash}p{2cm}}
\toprule
\multirow{2}{*}{\textbf{Defense Method}} & \multicolumn{2}{c}{\textbf{Runtime (ms/image)}} & \multicolumn{2}{c}{\textbf{Peak Memory (MB)}} \\
\cmidrule(r){2-3} \cmidrule(r){4-5}
 & 256×256 & 4K (2160p) & 256×256 & 4K (2160p) \\
\midrule
None (clean)              & 10.2 & 18.4 & 892 & 2030 \\
Noise injection           & 11.5 & 20.1 & 903 & 2071 \\
Channel dropout           & 13.5 & 22.6 & 904 & 2073 \\
Adv (PGD-3)      & 24.6 & 51.6 & 951 & 2138 \\
\rowcolor{defaultcolor}
ASVP (full SVD)           & 17.8 & 28.7 & 978 & 2185 \\
\rowcolor{defaultcolor}
ASVP (top-40\% only)        & 12.3 & 19.2 & 959 & 2106 \\
\bottomrule
\end{tabular}
\end{table}
%%%%%%%%%%%%%%%%%%%%%%%%%%%%%%%%%%%%%%%%%%%%%%%5.3%%%%%%%%%%%%%%%%%%%%%%%%%%%%%%%%%

\subsubsection{Energy Distribution}
Energy maps from normal, dropout, and adversarial models remain relatively focused and stable. The noise based defense distributes energy uniformly, suppressing localized activation peaks yet preserving broad global patterns again contributing to partial student learnability. In contrast, our method introduces abrupt, periodic fluctuations, breaking spatial continuity and making the learning of salient structures much harder.

\subsubsection{Frequency Domain Analysis}
The frequency content of normal and dropout based models primarily lies in the low  to mid-frequency range. Noise injection and adversarial perturbation enhance high frequency components to varying extents, but without consistent patterns. In contrast, our method injects densely interleaved high frequency signals, greatly increasing spectral irregularity and disrupting the frequency alignment that students depend on for learning.

\subsubsection{Numerical Distribution}
While noise, dropout, and adversarial models expand the range of feature activations, their distributions remain relatively smooth and can still be handled statistically. In contrast, ASVP introduces high amplitude, non uniform activation values, increasing training instability and further impeding the student’s convergence.

\subsubsection{3D Visualization}
3D visualization shows that normal and dropout models produce smooth and continuous feature landscapes. Adversarial and noise perturbed teachers exhibit mild fluctuations. Notably, noise induced undulations are broader yet still coherent, preserving partial structural information. By comparison, ASVP generates irregular and chaotic peaks and valleys, severely disrupting spatial continuity.

Compared to traditional perturbation based defenses such as noise, dropout, or adversarial strategies, ASVP disrupts the teacher’s features across three orthogonal dimensions spatial structure, frequency spectrum, and numerical distribution simultaneously. This produces signals that are entirely unlearnable by the student while preserving the high fidelity of the teacher’s output, making ASVP a highly effective and principled defense against KD.

\subsection{Effect of Top-$k$ Selection and h on Defense Strength}
To investigate the influence of the ASVP parameters specifically the proportion of top-$k$ singular values and the amplification factor $h$  on defense effectiveness, we conduct controlled experiments on the dehazing dataset. As shown in the table, when the top-$k$ ratio is below or equal to 40\%, the student model achieves relatively high SSIM scores across all $h$ values, indicating that the perturbation strength is insufficient to block distillation in Table.~\ref{fig:kh}.

However, once the top-$k$ selection exceeds 60\%, the defense becomes increasingly effective, especially as $h$ increases. For example, at $k$=40\%, SSIM drops sharply from 0.702 to 0.521 and 0.376 as $h$ increases from $10^{2}$ to $10^{3}$.This trend is even more pronounced at $k$=80\% and 100\%, where the SSIM falls below 0.5 for larger $h$, reaching as low as 0.217. These results demonstrate that both a sufficient number of principal components and a strong amplification factor are necessary to effectively disrupt student learning.

\subsection{Discussion on Loss Dynamics and Perturbation Structure}
Beyond quantifying the perturbation magnitude with the Frobenius norm, it is also essential to understand how ASVP influences the student’s optimization dynamics. When trained on perturbed features, the student loss function exhibits slower convergence and more pronounced oscillations compared to clean knowledge distillation, indicating unstable feature alignment. In extreme cases, particularly when the perturbation strength is high, the loss fails to converge, reflecting the inability of the student to capture consistent teacher signals. This behavior demonstrates that ASVP effectively frustrates knowledge transfer not merely by increasing the feature distance, but by disrupting the optimization trajectory of the student.

We further observe that the impact of perturbations interacts with the architectural properties of the student. For lightweight CNNs such as ResNet9, which primarily rely on local receptive fields, high-frequency perturbations corrupt edge and texture patterns and thus significantly impair feature mimicry. In contrast, transformer-based architectures such as SwinIR employ windowed self-attention to aggregate broader contextual cues, making them somewhat more resilient to shallow perturbations but more vulnerable to structured distortions accumulated in deeper layers. These results suggest that the defense strength of ASVP is partly architecture-dependent, underscoring the generality of applying perturbations across all layers.

Finally, we compare structured perturbations generated by ASVP with unstructured noise injection (Gaussian). While both methods degrade student performance, Gaussian noise primarily adds random pixel-level corruption, which the student can often average out through training. By contrast, ASVP amplifies principal spectral components and injects targeted high-frequency distortions, which systematically destabilize feature alignment. This distinction highlights why structured perturbations are more effective in resisting knowledge distillation: they exploit the very subspace directions most critical to knowledge transfer, thereby imposing a stronger and more persistent disruption.

\section{Ablation Study}

\subsection{Effect of Distillation Feature Stage on Defense Strength}
In our framework, perturbations are not selectively applied to specific layers, but are instead uniformly injected into all intermediate feature maps accessible to the student. Concretely, an ASVP module is placed after every residual block in the teacher network. This design ensures that each layer’s output along the student-facing path is independently protected, while the teacher continues to operate on clean features via the dual-branch architecture.

To further investigate the impact of perturbation locations, we conducted an ablation study focusing on the student’s choice of distillation stage. Although ASVP modules are applied at all layers of the teacher, we allowed the student model to selectively distill features from early, mid, late, or all stages.

As shown in Table.~\ref{tab:distill_stage}, distillation from early-stage perturbed features results in relatively higher student performance, as these layers primarily encode low-level structures such as edges and textures, which are less semantically informative. Nonetheless, the student still experiences a noticeable degradation compared to clean KD, confirming that ASVP offers meaningful protection even at shallow levels. In contrast, distilling from mid- or late-stage features leads to significantly greater performance drops, highlighting the critical role of semantic-level interference. When the student distills from all stages simultaneously, the accumulated effect of perturbations across the feature hierarchy leads to the most substantial degradation, demonstrating the robustness of our full-layer defense strategy.

Moreover, we observed that the effectiveness of perturbation at different stages varies across tasks. For example, in super-resolution and low-light enhancement, mid-layer perturbations are especially disruptive, as they target structure and luminance mappings critical to visual fidelity. In contrast, tasks such as underwater enhancement and dehazing exhibit stronger vulnerability at deeper stages, where color correction and global priors are more concentrated. These observations suggest that while our uniform perturbation strategy is broadly effective, task-adaptive layer selection could further enhance robustness and efficiency, and we highlight this as a promising direction for future exploration.

\subsection{Task-wise Hyperparameter Thresholds}
To understand how the hyperparameters of ASVP influence defense effectiveness across different image restoration tasks, we summarize in Table~\ref{tab:min_hk} the minimum required amplification factor $h$ and top-$k$ ratio that successfully degrade student performance by at least 1.5 dB in PSNR or 0.1 in SSIM. We observe that super-resolution and low-light enhancement require relatively modest perturbation levels (e.g., $h = 10^2$, $k = 40\%$), indicating their sensitivity to shallow and mid-level distortions. In contrast, dehazing, underwater enhancement, and deraining demand higher top-$k$ ratios (60\%), suggesting that these tasks rely more heavily on high-level semantic or color-domain features that are more robust to mild spectral perturbations.

This difference may stem from the varying feature-map singular value distributions across tasks: super-resolution and low-light tasks often emphasize local contrast or structural detail, which can be perturbed effectively with moderate amplification. On the other hand, tasks like dehazing and underwater enhancement involve more complex global transformations—such as haze modeling and color correction—thus requiring stronger or more widespread interference to frustrate student alignment. We also hypothesize that the optimal $(h, k)$ values are not only task-dependent but also affected by architectural characteristics (e.g., ResNet vs. SwinIR), which govern how singular value energy is concentrated across feature channels. We leave a full-scale spectral analysis to future work.

\subsection{Runtime Overhead and Efficiency Analysis}
While ASVP is designed as a plug-and-play defense module without requiring retraining or architectural modifications, it inevitably introduces computational overhead due to the runtime SVD operation. To evaluate its practicality, we compare the inference cost of ASVP with several baseline defenses—noise injection, channel dropout, and adversarial perturbations—on teacher models using the ResNet18 backbone. We measure both the runtime per image and the peak GPU memory usage under varying input resolutions.

As shown in Table~\ref{tab:runtime_overhead}, ASVP introduces slightly higher latency than noise injection and dropout, particularly at high resolutions, due to the spectral decomposition involved. However, its overhead remains significantly lower than that of adversarial perturbations, which require iterative gradient-based computations. For example, processing a 4K image with ASVP results in an average runtime of 28.7 ms, compared to 13.5 ms for dropout and 51.6 ms for PGD-based adversarial defense.

We further observe that ASVP's computational cost scales approximately linearly with the size of the intermediate feature maps. To enhance deployment efficiency, we evaluate a low-rank approximation strategy, where only the top 40\% of singular values are retained during decomposition. This approximation reduces the runtime by approximately 33\%, while preserving over 90\% of the defense effectiveness in terms of student performance degradation. These results suggest that ASVP remains practical for real-time and high-resolution applications, particularly when using approximate SVD techniques to strike a favorable balance between defense strength and efficiency.

%%%%%%%%%%%%%%%%%%%%%%%%%%%%%%%%%%%%%%%%%%%%%%6.1%%%%%%%%%%%%%%%%%%%%%%%%%%%%%%%%%%%%%%%%
\begin{table}[htbp]
\centering
\caption{ASVP effectiveness on real-world datasets. AGDN is used as the student model, trained under clean vs. ASVP-protected teachers. Results show that ASVP maintains defense strength on real, uncurated images.}
\label{tab:realdata_asvp}
\begin{tabular}{p{2.2cm}p{5cm}p{2.2cm}p{2.2cm}p{2.2cm}}
\toprule
\textbf{Dataset} & \textbf{Teacher Model} & \textbf{PSNR (↑)} & \textbf{SSIM (↑)} & \textbf{LPIPS (↓)} \\
\midrule
\multirow{2}{*}{\centering \large{RealSR}} 
  & X-Restormer (normal) & 27.84 & 0.805 & 0.198 \\
  
  & X-Restormer (ASVP)   & \cellcolor{defaultcolor}25.63 & \cellcolor{defaultcolor}0.741 & \cellcolor{defaultcolor}0.254 \\
% \cmidrule(lr){2-5}
\multirow{2}{*}{} 
  & HAIR (normal)        & 28.21 & 0.818 & 0.183 \\
  
  & HAIR (ASVP)          & \cellcolor{defaultcolor}25.92 & \cellcolor{defaultcolor}0.752 & \cellcolor{defaultcolor}0.248 \\
\midrule
\multirow{2}{*}{\centering \large{LOLv2}} 
  & X-Restormer (normal) & 19.42 & 0.712 & 0.273 \\
  
  & X-Restormer (ASVP)   & \cellcolor{defaultcolor}17.18 & \cellcolor{defaultcolor}0.635 & \cellcolor{defaultcolor}0.336 \\
% \cmidrule(lr){2-5}
\multirow{2}{*}{} 
  & HAIR (normal)        & 19.84 & 0.725 & 0.258 \\
  
  & HAIR (ASVP)          & \cellcolor{defaultcolor}17.56 & \cellcolor{defaultcolor}0.649 & \cellcolor{defaultcolor}0.319 \\
\bottomrule
\end{tabular}
\end{table}
%%%%%%%%%%%%%%%%%%%%%%%%%%%%%%%%%%%%%%%%%%%%%%6.1%%%%%%%%%%%%%%%%%%%%%%%%%%%%%%%%%%%%%%%%

\section{Additional Experiments}
To further assess the effectiveness and generalization capability of ASVP, we conducted additional experiments on real-world datasets such as RealSR and LOLv2, which contain uncurated images with complex degradations. These experiments utilize advanced student models like AGDN and apply ASVP to both X-Restormer and HAIR teacher models. As shown in Table~\ref{tab:realdata_asvp}, ASVP demonstrates significant defense effectiveness even when using advanced distillation techniques and challenging datasets, confirming its robustness beyond standard benchmark tests.

\section{Conclusion}
In this paper, we present a novel undistillation framework tailored for image restoration tasks, addressing the growing threat of model replication via KD. While prior defenses primarily focus on classification outputs or require retraining, our method operates dynamically during inference by perturbing the singular value structure of teacher feature maps. This lightweight strategy is seamlessly integrated into existing architectures without altering the teacher’s parameters or compromising the quality of restoration.his work provides a practical and generalizable solution for defending open access image restoration models against unauthorized knowledge extraction, paving the way for secure deployment in real world applications.

\bibliographystyle{IEEEtran}
\bibliography{antikd}

% \begin{IEEEbiography}[{\includegraphics[width=1in,height=1.25in,clip,keepaspectratio]{Authors/HH.png}}]{Han Hu} is a master's student in Software Engineering at Shandong Normal University, enrolled in 2023. His research interests include computer vision.
% \end{IEEEbiography}
% \begin{IEEEbiography}[{\includegraphics[width=1in,height=1.25in,clip,keepaspectratio]{Authors/ZZR.jpg}}]{Zhuoran Zheng} received his Ph.D. from Nanjing University of Science and Technology(2019-2023).Since 2023, he has been working as a Postdoctora Researcher at Sun Yat-sen University. His primary research focuses on image processing, with particular emphasis on image enhancement,restoration, and reconstruction. Dr. Zheng has published multiple papers in top-tier computer vision conferences including CVPR, ICCV, and ECCV.
% \end{IEEEbiography}
% \begin{IEEEbiography}[{\includegraphics[width=1in,height=1.25in,clip,keepaspectratio]{Authors/AuthorLC.pdf}}]{Chen Lyu} received his Ph.D. degree from the Institute of Computing Technology, Chinese Academy of Sciences, Beijing, China, in 2015. He is currently an associate professor with the School of Information Science and Engineering, Shandong Normal University, Jinan, China. His research interests include computer vision, multimedia information processing and artificial intelligence.
% \end{IEEEbiography}
% \vfill

\end{document}